\documentclass{article}

\usepackage{graphicx}
\usepackage[caption=false]{subfig}
\usepackage{float}
\usepackage{stfloats}
\usepackage{amsmath}
\usepackage{amssymb}
\usepackage{physics}  % some symbols such as abs and norm

\usepackage[numbers]{natbib}

\usepackage{color}

\usepackage{mathtools}

\usepackage{etoolbox}
\usepackage[algo2e, ruled, noend, commentsnumbered]{algorithm2e}
\SetAlCapNameFnt{\scriptsize}
\SetAlgorithmName{Alg.}{}{}

\makeatletter
\patchcmd{\@algocf@start}% <cmd>
  {-1.5em}% <search>
  {0pt}% <replace>
  {}{}% <success><failure>
\makeatother

\SetAlCapHSkip{0em}
\SetInd{0.4em}{0.4em}

\SetKw{KwWhile}{while}
\SetKwArray{prev}{prev}
\SetKwArray{curr}{curr}
\SetKwFunction{swap}{swap}
\SetKwFunction{co}{co}
\SetKwFunction{li}{li}
\SetKwFunction{kwmin}{min}
\SetKwData{nextstart}{next\_start}
\SetKwData{pruningpoint}{pruning\_point}
\SetKwData{nextpruningpoint}{next\_pruning\_point}
\SetKwData{ub}{ub}
\SetKwData{cutoff}{cutoff}
\SetKwData{lb}{lb}
\SetKwData{jStart}{jStart}
\SetKwData{jStop}{jStop}
\SetKwData{lb}{lb}

\usepackage{hyperref}

\newcommand\R{\mathrm}
\newcommand\MC{\mathcal}

\DeclareMathOperator{\DTW}{DTW}
\DeclareMathOperator{\CDTW}{CDTW}
\DeclareMathOperator{\WDTW}{WDTW}
\DeclareMathOperator{\ERP}{ERP}
\DeclareMathOperator{\MSM}{MSM}
\DeclareMathOperator{\TWE}{TWE}

\DeclareMathOperator{\cost}{cost}

\begin{document}

\title{Early Abandoning and Pruning for Elastic Distances including Dynamic Time Warping\thanks{This research has been supported by Australian Research Council grant DP210100072.}
}

\author{Matthieu Herrmann,  Geoffrey I. Webb \\
  {\normalsize Monash University, Australia} \\
  {\footnotesize \texttt{\{matthieu.herrmann,geoff.webb\}@monash.edu}}
}

\date{3 June 2021}

\maketitle

\begin{abstract}
  Nearest neighbor search under elastic distances is a key tool for time series analysis, supporting many applications.
  However, straightforward implementations of distances require $O(n^2)$ space and time complexities,
  preventing these applications from scaling to long series.
  Much work has been devoted to speeding up the NN search process,
  mostly with the development of lower bounds,
  allowing to avoid costly distance computations when a given threshold is exceeded.
  This threshold, provided by the similarity search process,
  also allows to early abandon the computation of a distance itself.
  Another approach, is to prune parts of the computation.
  All these techniques are orthogonal to each other.
  In this work, we develop a new generic strategy, ``EAPruned'',
  that tightly integrates pruning with early abandoning.
  We apply it to six elastic distance measures: DTW, CDTW, WDTW, ERP, MSM and TWE,
  showing substantial speedup in NN search applications.
  Pruning alone also shows substantial speedup for some distances,
  benefiting applications beyond the scope of NN search (e.g. requiring all pairwise distances),
  and hence where early abandoning is not applicable.
  We~release our implementation as part of a new C++ library for time series classification,
  along with easy to use Python/Numpy bindings.
\end{abstract}

% --- --- --- --- --- --- --- --- --- --- --- --- --- --- --- --- --- --- --- --- --- --- --- --- --- --- --- --- --- ---
% --- --- --- --- --- --- --- --- --- --- --- --- --- --- --- --- --- --- --- --- --- --- --- --- --- --- --- --- --- ---
\section{\label{intro}Introduction}
Nearest neighbor (NN) search under elastic distances is a major tool in time series analysis, supporting many applications, including
classification \citep{linesTimeSeriesClassification2015},
sub-sequence search \citep{silvaSpeedingSimilaritySearch2018},
regression \citep{RN3458},
clustering \citep{petitjeanGlobalAveragingMethod2011},
and outlier detection \citep{benkabou2018unsupervised}.

Unfortunately, elastic distances have  quadratic time complexity with respect to length,
incurring costly computation and preventing many applications from scaling to long series.
This has mostly been addressed through lower bounding,
which seeks to identify cases where the quadratic distance computation can be avoided (see Section~\ref{sec:bg-sslb}).
Two main strategies have been developed for directly speeding up distance computations:
``pruning'' and ``early abandoning'' (see Section~\ref{sec:bg-pruning_and_ea}).
In this paper we develop a new algorithm, ``EAPruned'',
that tightly integrates both pruning and early abandoning, thereby substantially reducing computation and often rendering lower bounding superfluous.

EAPruned is a generic strategy, applicable to a broad class of elastic distances.
We investigate the effectiveness of our approach with six key elastic distance measures:
DTW, CDTW, WDTW, ERP, MSM and TWE.
To enable fair comparison, we implemented the key alternative algorithms for these six measures in C++
and compared their run times using NN classification over the UCR archive
(in its 85 univariate datasets version).
We show that EAPruned offers significant speedups:
from $3.93$ to $39.23$ times faster than simple implementations,
and from $1.85$ to $8.44$ times faster than implementations
with the usual early abandoning scheme.
We also show that our algorithm remains effective in pruning-only uses,
i.e. when early abandoning is not applicable
We then show, in the cases of DTW and CDTW, that lower bounding is complementary to EAPruned.
Finally, we show that our algorithm is also effective in settings beyond NN classification,
with an application to sub-sequence search.

The rest of this paper is organised as follows.
Section~\ref{sec:bg} introduces the context of this work and presents the six distances.
The related work is presented in Section~\ref{sec:relwork}, and EAPruned itself is described in Section~\ref{sec:eapruned}.
We then present our experimental results in Section~\ref{sec:experiments},
and conclude in Section~\ref{sec:5}.

% --- --- --- --- --- --- --- --- --- --- --- --- --- --- --- --- --- --- --- --- --- --- --- --- --- --- --- --- --- ---
% --- --- --- --- --- --- --- --- --- --- --- --- --- --- --- --- --- --- --- --- --- --- --- --- --- --- --- --- --- ---
\section{\label{sec:bg}Background}
NN classification has historically been the workhorse of time series classification, 
which led to the development of various elastic distances.
The most widely used of these is the Dynamic Time Warping (DTW) distance,
introduced in 1971 by \cite{SakoeChiba71}
along with its constrained variant CDTW \citep{sakoeDynamicProgrammingAlgorithm1978}.
Guided by the UCR time series archive~\citep{dauUCRTimeSeries2019},
time series classification has made enormous progress in the past decade,
including the rise of ``ensemble classifiers'',
i.e. classifiers combining an ensemble of other classifiers.
The ``Elastic Ensemble''  (``EE'', see~\cite{linesTimeSeriesClassification2015}), introduced in 2015,
was one of the first classifiers to be consistently more accurate than NN-DTW over a wide variety of tasks.
EE combines eleven NN classifiers based on eleven previously developed elastic distances
(see Section~\ref{sec:bg:ED}).

Most elastic distances have parameters.
At train time, EE fine tunes these parameters using cross validation.
At test time, the query's class is determined through a majority vote between the component classifiers.
``Proximity Forest'' (``PF'', see~\cite{lucasProximityForestEffective2019})
uses the same set of NN classifiers as EE,
but deploys them within an ensemble of random decision trees,
for which splits are determined by distance to exemplars of each class.
Both EE and PF solely work in the time domain,
leading to poor accuracies when discriminant features are in other domains.
This was addressed by their respective evolution into ``HIVE-COTE''
(``HC'', see~\cite{linesTimeSeriesClassification2018})
and ``TSCHIEF'' (see~\cite{shifazTSCHIEFScalableAccurate2020}),
combining more classifiers working in different domains
(interval, shapelets, dictionary and spectral, see~\cite{linesTimeSeriesClassification2018} for an overview).

While EE provided a qualitative jump in accuracy,
it also required a quantitative jump in computation time.
For example, \cite{tanFastEEFastEnsembles2020}
reports 17 days to learn and classify the UCR archive's ElectricDevice benchmark.
Indeed, given series of length $L$, naive elastic distance algorithms have $O(L^2)$ space and time complexities.
This is only compounded by an extensive search for the best 
parametrization through leave-one-out cross validation.
For a training set of $N$ times series, searching among $M$ parameters ($M=100$ for EE),
EE has a training time complexity in $O(M.N^2.L^2)$.
HIVE-COTE not only expands on EE, it actually embeds it as a component -- or did.
Recently, EE was dropped from HIVE-COTE (see~\cite{bagnallTaleTwoToolkits2020})
because of its computational cost.
This caused an average drop of $0.6\%$ in accuracy,
and beyond a $5\%$ drop on some datasets (tested on the UCR archive).
The authors considered that this loss in accuracy was a necessary price to pay for the resulting speedup
(the authors only report that the new version is ``significantly faster''). 
The implication is that if NN classifiers can be sped up sufficiently,
it will be possible for EE to rejoin HIVE-COTE,
resulting in a substantial further lift in accuracy in what is currently the most accurate time series classifier.

Speeding up the computation of NN search under elastic distances is also of interest
for several other reasons.
First, TS-Chief~\citep{shifazTSCHIEFScalableAccurate2020},
which is another state-of-the-art classifier, still relies on NN classifiers.
Second, this will benefit other applications relying on NN search, such as
clustering~\citep{petitjeanGlobalAveragingMethod2011},
outlier detection~\citep{benkabou2018unsupervised},
regression \citep{RN3458},
and sub-sequence search~\citep{silvaSpeedingSimilaritySearch2018}.
Third, a disproportionate amount of research has been spent on DTW and CDTW compared to other distances,
which is reinforced by the lack of efficient lower bounds for their alternatives.
Our algorithm provides substantial speed-up for all six presented distances,
and for any further distance that follows a similar structure (see~Section~\ref{sec:bg}).

In this paper, we will only consider univariate time series,
although EAPruned is also applicable to multivariate series.
We denote series by $Q$ (for a query), $C$ (for a candidate), $S$, and $T$.
Their length is denoted by $L$, using subscript such as $L_S$ to disambiguate the lengths of different series.
Subscripts $C_i$ are used to distinguish multiple candidates.
The elements $s_1, s_2,\dots{}s_{L_S}$ are the elements of the series $S=(s_1, s_2, \dots s_{L_S})$.
The element $s_i$ is the $i$-th element of $S$ with $1\leq{}i\leq{}L_S$.

\subsection{\label{sec:bg-sslb}Similarity Search and Lower Bounding}
Nearest neighbor search is a branch of similarity search,
i.e. a search solely relying on the similarity between any pair of objects.
It is a common application of elastic distances,
and we use it to demonstrate the efficacy of our algorithm.
Given a query $Q$ and a set of $n$ candidates $\MC{C} = \{C_1, C_2, \dots C_n\}$,
the nearest neighbor of $Q$ under a distance $D$ is a candidate $C_\text{nn}$
with $d_\text{nn}=D(Q, C_\text{nn})$ such that $\forall{C\in{\MC{C}}}, d_\text{nn}\leq{D(Q, C)}$.
Algorithm~\ref{alg:nnsearch} presents this process.

\begin{center}
\begin{minipage}[t]{0.48\linewidth}
  \vspace{0pt}  
  \begin{algorithm2e}[H]
    \small
    \SetAlgoLined
    \LinesNumbered
    \DontPrintSemicolon
    \caption{\label{alg:nnsearch}NN search}
    $(d_\text{nn}, C_\text{nn}) \leftarrow (\infty, \emptyset)$\;
    \For{$C \in \MC{C}$}{
        $d \leftarrow D(Q, C)$ \;
        \lIf{$d<d_\text{nn}$}{ $(d_\text{nn}, C_\text{nn}) \leftarrow (d, C)$ }
    }
    \Return $(d_\text{nn}, C_\text{nn})$\;
  \end{algorithm2e}
\end{minipage}\hspace{0.02\linewidth}%
\begin{minipage}[t]{0.50\linewidth}
  \vspace{0pt}
  \begin{algorithm2e}[H]
    \small \SetAlgoLined \LinesNumbered \DontPrintSemicolon
    \caption{\label{alg:nnsearchLB}Lower bounded NN search}
    $(d_\text{nn}, C_\text{nn}) \leftarrow (\infty, \emptyset)$\;
    \For{$C \in \MC{C}$}{
        \If{$\mbox{LB}(Q, C) < d_\text{nn}$}{
            $d \leftarrow D(Q, C)$ \;
            \lIf{$d<d_\text{nn}$}{ $(d_\text{nn}, C_\text{nn}) \leftarrow (d, C)$ }
    }}
    \Return $(d_\text{nn}, C_\text{nn})$\;
  \end{algorithm2e}
\end{minipage}
\end{center}

Note that at any point in time during the execution of Algorithm~\ref{alg:nnsearch},
$d_\text{nn}$ is a monotonously decreasing upper bound on the end result.
We will refer to this upper bound as the ``cut-off''.
Any candidate $C$ is discarded if $D(Q,C)\geq d_\text{nn}$.
Lower bounding exploits this fact to speed up the NN search (Algorithm~\ref{alg:nnsearchLB}).
A lower bound $\mbox{LB}$ of a distance $D$ is an approximation of $D$
such that $\mbox{LB}(Q,C)\leq{}D(Q,C)$.
$\mbox{LB}(Q,C)\geq d_\text{nn} \models D(Q, C)\geq d_\text{nn}$,
allowing us to skip its computation as its result will be discarded.

An ideal lower bound is fast (usually in $O(L)$, when distances are in $O(L^2)$)
and tight (as close as possible to the actual distance).
Lower bounding is shown to significantly speedup NN search in several domains
\citep{keoghExactIndexingDynamic2005a, tanFastEEFastEnsembles2020, rakthanmanonSearchingMiningTrillions2012}.
Lower bounds have mainly been developed for DTW and CDTW, two widely used examples being LB Kim \citep{sang-wookkimIndexbasedApproachSimilarity2001} and LB Keogh
\citep{keoghExactIndexingDynamic2005a}.
They also exist for other elastic distances \citep{tanFastEEFastEnsembles2020},
and remain an active field of research \citep{WEBB2021107895}.

Another branch of similarity search is range queries,
in which we want to find all candidates within a given distance of a given exemplar.
Range queries benefit from both lower bounding and early abandoning (see Section~\ref{sec:bg-pruning_and_ea}),
quickly discarding candidates beyond a cut-off, which in this case is the maximum requested distance.

\subsection{\label{sec:bg-pruning_and_ea}Pruning and Early Abandoning}
``Pruning'' and ``early abandoning'' are generic concepts.
``Pruning'' refers to identifying and avoiding unproductive operations.
``Early abandoning'' refers to abandoning the whole computation as soon as
it can be established through an ``abandoning criterion'' that an exact result is not required.
In this paper, we will say that the distance computation algorithms are pruned and early abandoned,
meaning that they support pruning and early abandoning.
When describing a computation, we will also say that some operations are pruned or early abandoned,
meaning that the actual act of pruning or early abandoning is taking place.

%A pruned distance computation algorithm always returns an exact result.
%which may abandon before determining the exact result,
%and requires an abandoning criterion as an extra parameter.
A pruned distance computation always returns an exact similarity score,
unlike the case for an early abandoned one,
which may abandon before determining the similarity score.
Moreover, early abandoning requires an abandoning criterion as an extra parameter.
Hence, early abandoned distances require special support from their caller
whereas pruned distances can be used in place of their straightforward counterparts.
NN search easily provides this support:
the abandoning criterion is the same cut-off used by lower bounding (see Section~\ref{sec:bg-sslb}),
and signaling early abandoning by returning $\infty$ is immediately compatible with the NN search algorithm.

The usual way DTW and other elastic distance computation algorithms
have been early abandoned until now~\citep{mueenExtractingOptimalPerformance2016}
is by monitoring the minimal cost of the distance at any point on the current boundary of the computed paths, and abandoning when it exceeds the cut-off.
This approach is presented in Algorithm~\ref{alg:CommonWEA}, Section~\ref{sec:Common}.
Of the numerous elastic distance measures, to the best of our knowledge pruning has previously only be developed for DTW (see Section~\ref{sec:relwork})
and is explained in Section~\ref{sec:eapruned}.

\subsection{\label{sec:bg:ED}Presentation of Elastic Distances}
Many elastic distances share a common form,
captured by Equations~\ref{eq:Common:corner} to~\ref{eq:Common:main}.

\begin{subequations}\label{eq:Common}
 \begin{align}
    M_{D(S,T)}(0,0) &= 0 \label{eq:Common:corner}\\
    M_{D(S,T)}(i,0) &= \R{InitVBorder}
    \quad\text{with}\quad{}M_{D(S,T)}(i,0)\leq{}M_{D(S,T)}(i+1,0) \label{eq:Common:vborder}\\
    M_{D(S,T)}(0,j) &= \R{InitHBorder}
    \quad\text{with}\quad{}M_{D(S,T)}(0, i)\leq{}M_{D(S,T)}(0, i+1)\label{eq:Common:hborder}\\
    M_{D(S,T)}(i,j) &= \min\left\{
    \begin{aligned}
        &M_{D(S,T)}(i-1, j-1) + \R{Canonical}  \\
        &M_{D(S,T)}(i-1, j) + \R{AlternateRow} \\
        &M_{D(S,T)}(i, j-1) + \R{AlternateColumn}
        \end{aligned}
    \right. \label{eq:Common:main}
 \end{align}
\end{subequations}

Elastic distances following this form include DTW, CDTW, WDTW, ERP, MSM, and TWE, which,
when coupled with taking the first derivative of the series \citep{keoghDerivativeDynamicTime2001}
before applying these distances (leading to DDTW, DCDTW, and DWDTW),
%transforms applied to the series before applying these distances%
account for nine of the 11 distance measures used by both EE and PF.
The 2 remaining distances, LCSS and SQED, do not share the same form,
hence do not benefit from our EAPruned approach.

An elastic distance $D$ computes an optimal, minimal, alignment cost $D(S,T)$
between two series $S$ and $T$ by minimizing the cumulative cost of aligning their individual points.
A ``cost matrix'' $M_{D(S,T)}$ is a 0-indexed matrix with dimension ${(1+L_S, 1+L_T)}$
used to carry this computation.
A cell $M_{D(S,T)}(i,j)$ represents the minimal cumulative cost of aligning the fist $i$ points of $S$
with the first $j$ points of $T$.
It follows that the cell $M_{D(S,T)}(L_S, L_T)$ holds the cost $D(S, T)$.
Equations~\ref{eq:Common:corner} to~\ref{eq:Common:main} give a generic form to compute a cost matrix.
An example of a cost matrix for DTW and the corresponding individual point alignments
are shown Figure~\ref{fig:DTW}.
An alignment between the points $s_i$ and $t_j$ is represented by the couple $(i,j)$.
In the cost matrix $M_{D(S,T)}$, the successive alignments ${(1,1), \dots (i,j) \dots (L_S, L_T)}$
form a path called the ``warping path'' (cells with black borders in Figure~\ref{fig:DTW:Matrix}).
Conversely, reading the warping path gives the successive alignment of the individual points.
See how the horizontal section of the warping path line 5 in Figure~\ref{fig:DTW:Matrix}
corresponds to the fifth point of $S$ being aligned thrice in Figure~\ref{fig:DTW:Alignments}.

Equations~\ref{eq:Common:corner} to~\ref{eq:Common:hborder} initialise the borders.
They are monotonously increasing, i.e. $M_{D(S,T)}(i, 0)\leq{}M_{D(S,T)}(i+1, 0)$.
Equation~\ref{eq:Common:main} computes the value of every other cell $(i,j)$ by taking
the minimum of three alignment costs plus their respective dependency.
The ``Canonical'' alignment depends on the top left diagonal cell $(i-1, j-1)$ and represents
an alignment between two new points.
For example, the cell $(3,2)$ in Figure~\ref{fig:DTW:Matrix}
represents $t_2$ being aligned with $s_3$ in Figure~\ref{fig:DTW:Alignments}.
The ``AlternateRow'' alignment depends on the top cell $(i-1, j)$
and represents an alignment between a new point along the lines,
and reusing the last aligned point along the columns.
For example, the cell $(4,2)$ in Figure~\ref{fig:DTW:Matrix}
represents $t_2$ being aligned with $s_4$ in Figure~\ref{fig:DTW:Alignments},
when $t_2$ was already aligned with $s_3$.
The ``AlternateColumn'' alignment depends on the left cell $(i, j-1)$,
and is the symmetric of the AlternateRow alignment for the column.
Alternate alignments happen either because the canonical alignment is too expensive,
or because the series have differing lengths.
In several instances, the canonical and alternate costs depend on a ``$\cost$'' function
between $s_i$ and $t_j$.
It usually is the squared Euclidean distance or the L1 norm, but other norms are acceptable.

These generic equations give us the following guarantees.
\begin{enumerate}
    \item The optimal alignment cost is computed.
        Note that several optimal warping paths with the same minimal cost may exist.
    \item Series extremities are aligned with each other ($(1,1)$ and $(L_S, L_T)$).
        A valid warping path starts from the top left cell and reaches the bottom right cell.
    \item The optimal warping path is continuous and monotonous,
    i.e.~for two of its consecutive cells ${(i_1, j_1) \neq (i_2, j_2)}$,
    we have ${i_1\leq{i_2}\leq{i_1+1}}$ and ${j_1\leq{j_2}\leq{j_1+1}}$.
    Graphically, no alignment (dotted line in Figure~\ref{fig:DTW:Alignments}) crosses each other.
\end{enumerate}

\subsubsection{Dynamic Time Warping}

\begin{figure}[t]
  \centering
  \subfloat[\label{fig:DTW:Alignments}
  $\DTW(S,T)$ alignments with costs.]{%
    \includegraphics[width=0.475\linewidth]{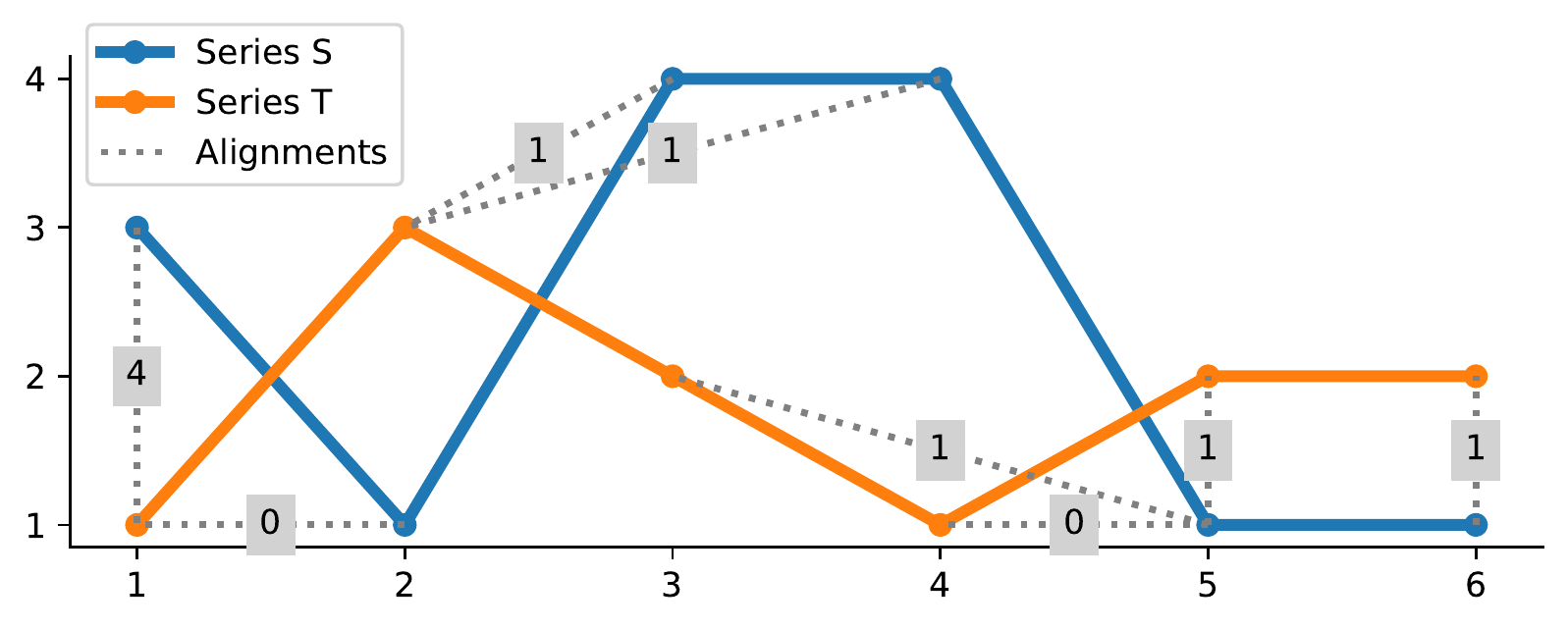}
  }
  \hspace{\fill}
  \subfloat[\label{fig:DTW:Matrix}
  $M_{\DTW}(S,T)$ cost matrix and warping path.
  ]{%trim = left bottom right top
    \includegraphics[trim=0 35 0 0, clip,width=0.475\linewidth]{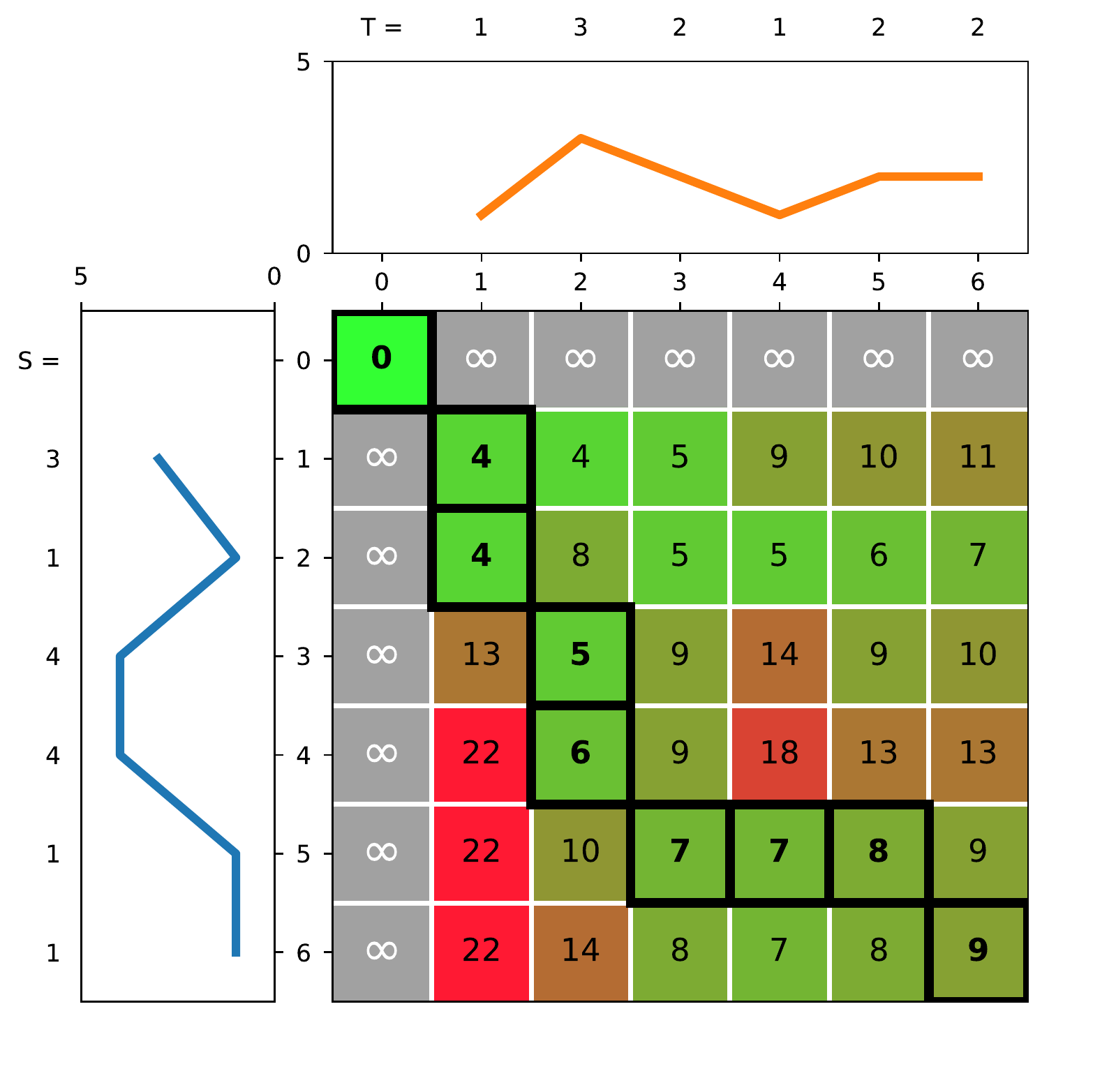}
  }%
  \vspace*{-3pt}\caption{\label{fig:DTW}
    $M_{\DTW(S,T)}$ with warping path and alignments between the series
    \mbox{$S = (3, 1, 4, 4, 1, 1)$} and \mbox{$T = (1, 3, 2, 1, 2, 2)$}.
    We have $\DTW(S,T)=M_{\DTW(S,T)}(6,6)=9$.
  }
\end{figure}

The Dynamic Time Warping (DTW, see Equations~\ref{eq:DTW:corner} to \ref{eq:DTW:main})
distance was first introduced %in 1971 by Sakoe and Chiba 
as a speech recognition tool
\citep{SakoeChiba71}.
Compared to the classic Euclidean distance, DTW handles distortion and disparate lengths.
Figure~\ref{fig:DTW} illustrates how DTW aligns two series $S$ and $T$,
along with the corresponding warping in the $M_{\DTW}(S,T)$ cost matrix.

\begin{subequations}\label{eq:DTW}
 \begin{align}
    M_{\DTW}(0,0) &= 0       \label{eq:DTW:corner}\\
    M_{\DTW}(i,0) &= +\infty \label{eq:DTW:vborder}\\
    M_{\DTW}(0,j) &= +\infty \label{eq:DTW:hborder}\\
    M_{\DTW}(i,j) &= \cost(s_i, t_j) + \min\left\{
    \begin{aligned}
        &M_{\DTW}(i-1, j-1) \\
        &M_{\DTW}(i-1, j) \\
        &M_{\DTW}(i, j-1)
    \end{aligned}
    \right. \label{eq:DTW:main}
 \end{align}
\end{subequations}

\subsubsection{\label{sec:CDTW}Constrained DTW}
\begin{figure}
  \centering
  \subfloat[\label{fig:CDTW:ok}CDTW cost matrix with $w=1$ for $S$ and $T$ of equal length]{%
    \includegraphics[trim=0 25 0 0,clip,width=0.475\linewidth]{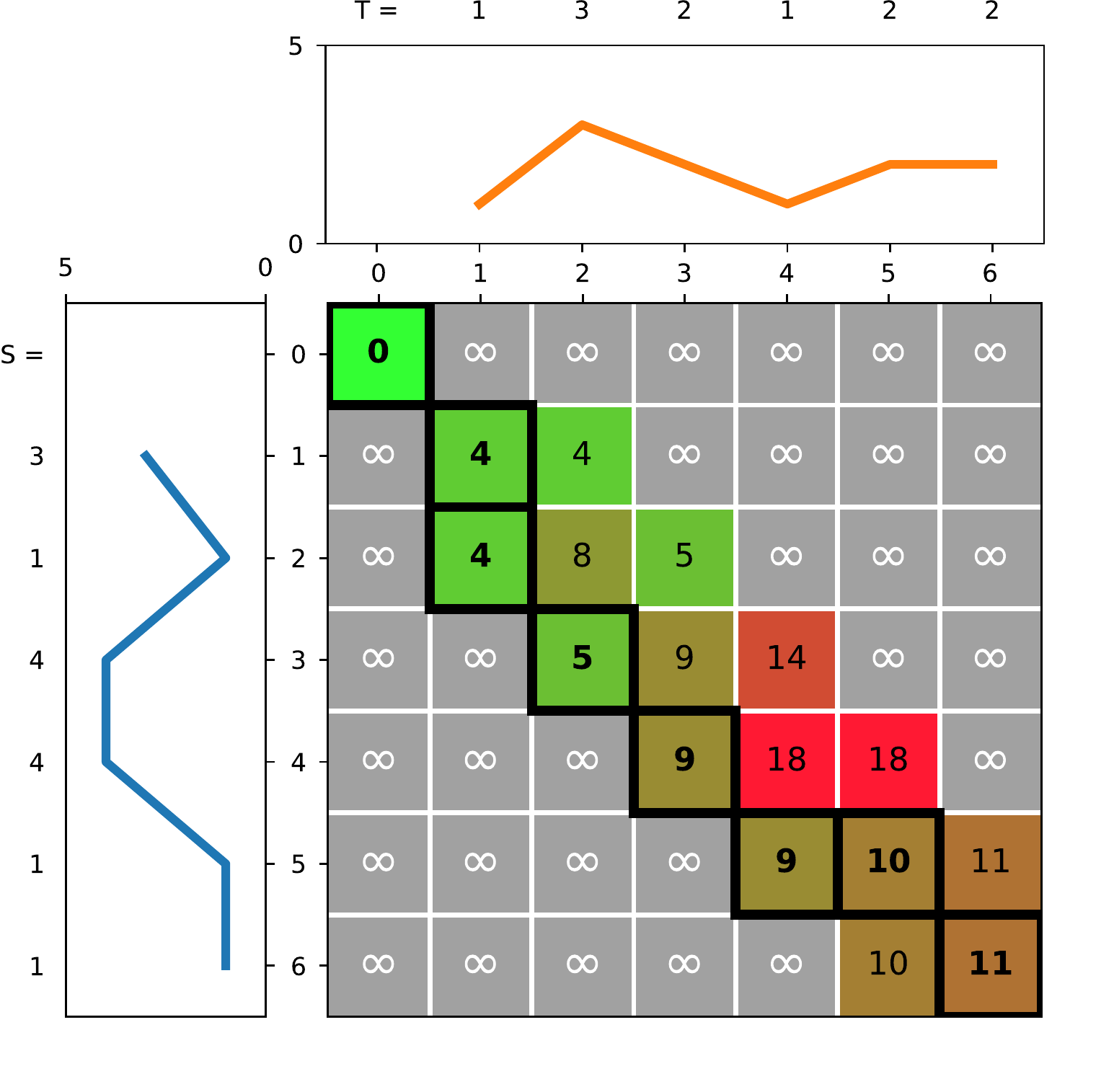}
  }%
  \hspace{\fill} 
  \subfloat[\label{fig:CDTW:notok}CDTW cost matrix with $w=1$ for $S$ and $U$ of disparate lengths.
  ]{%
    \includegraphics[trim=0 25 0 0, clip,width=0.475\linewidth]{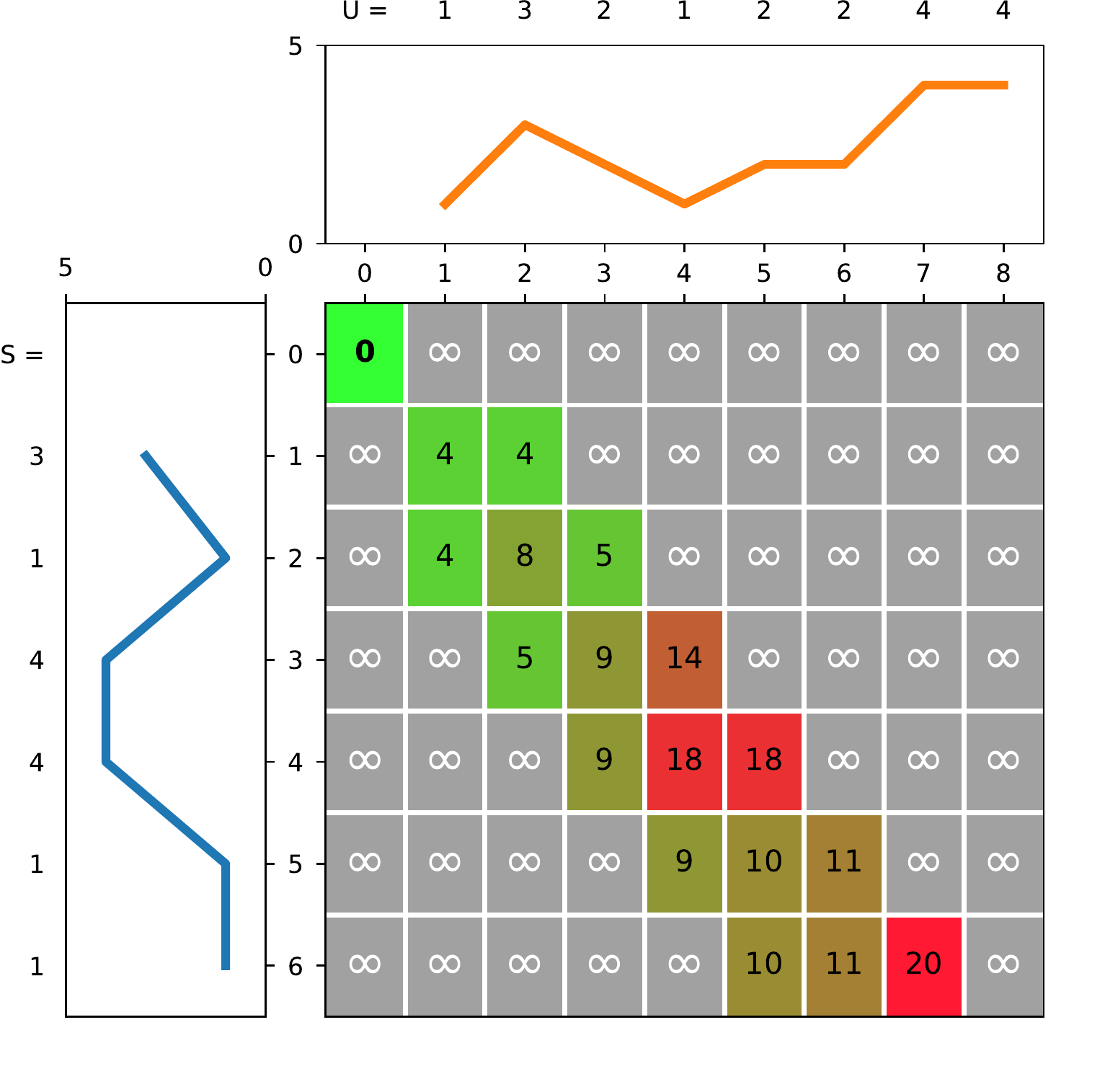}
  }%
  \vspace*{-4pt}\caption{\label{fig:CDTW} Example of CDTW cost matrices with a window of $1$.
  In the second case, the window it too small to allow an alignment between the last two points at $(6,8)$.
  }
\end{figure}

In CDTW, the ``constrained'' variant of DTW, 
the warping path is restricted to a subarea of the matrix.
Different constraints exist \citep{sakoeDynamicProgrammingAlgorithm1978, itakuraMinimumPredictionResidual1975}.
We focus on the popular Sakoe-Chiba band \citep{sakoeDynamicProgrammingAlgorithm1978},
also known as ``Warping Window'' (or ``window'' for short),
which also appears in ERP (Section~\ref{sec:ERP}).
The window is a parameter $w$ controlling how far the warping path can deviate from the diagonal.
Given a matrix $M$, a line index $1\leq{}l\leq{}L$ and a column index $1\leq{}c\leq{}L$,
we have $\abs{l-c}\leq{}w$.
For example with $w=1$,
the warping path can only step one cell away from each side of the diagonal (Figure~\ref{fig:CDTW:ok}).

A window of $0$ is akin to the squared Euclidean distance
(actually is, if the ``$\cost$'' function between point also is),
while a window of $L$ is equivalent to DTW (no constraint). 
With a correctly set window, NN-CDTW can achieve better accuracy than NN-DTW by preventing spurious alignments.
It is also faster to compute than DTW, as cells beyond the window are ignored.
However, the window parameter must be set.
Finally, not all window size are valid: if the series are of disparate lengths,
a window can be too small to allow any alignment (Figure~\ref{fig:CDTW:notok}).

\subsubsection{Weighted DTW}
Weighted Dynamic Time Warping (WDTW, see~\cite{jeongWeightedDynamicTime2011})
imposes a soft constraint on the warping path.
The cells $(l,c)$ of the $M_{\WDTW}$ cost matrix are weighted
according to their distance to the diagonal $d=\abs{l-c}$.
A large weight decreases the chances of a cell to be on an optimal path.
The weight is computed according to Equation~\ref{eq:WDTW}.
The parameter $g$ controls the penalization,
and usually lies within $0.01$ -- $0.6$~\citep{jeongWeightedDynamicTime2011}.
\begin{equation} \label{eq:WDTW}
    w(d)=\frac{1}{1+\exp^{-g\times{}(d-L/2)}}
\end{equation}

\subsubsection{\label{sec:ERP}Edit Distance with Real Penalty}
The Edit distance with Real Penalty (ERP, see Equations~\ref{eq:ERP:corner} to \ref{eq:ERP:main})
is a metric designed as a L1-norm supporting local time shifting,
or alternatively as a DTW variant fulfilling the triangular inequality~\citep{chenMarriageLpnormsEdit2004}.
It is parameterized by a warping window (see Section~\ref{sec:CDTW}) and a ``gap value'' $g$.
Compared to DTW, which reuses a previous point to compute an alternate cost
(e.g. in Figure~\ref{fig:DTW}, the point 5 of S is used thrice),
ERP computes an alternate cost based on $g$ (Equation~\ref{eq:ERP:main}).
This property allows to recover the triangular inequality absent from DTW~\citep{chenMarriageLpnormsEdit2004}.
ERP requires its borders to be computed (Equations~\ref{eq:ERP:vborder} and~\ref{eq:ERP:hborder}).

\begin{subequations}\label{eq:ERP}
 \begin{align}
    M_{\ERP}(0,0) &= 0       \label{eq:ERP:corner}\\
    M_{\ERP}(i,0) &= M_{\ERP}(i-1, 0) + \cost(s_i, g) \label{eq:ERP:vborder}\\
    M_{\ERP}(0,j) &= M_{\ERP}(0, j-1) + \cost(g, t_j) \label{eq:ERP:hborder}\\
    M_{\ERP}(i,j) &= \min\left\{
    \begin{aligned}
        &M_{\ERP}(i-1, j-1) + \cost(s_i, t_j) \\
        &M_{\ERP}(i-1, j) + \cost(s_i, g)\\
        &M_{\ERP}(i, j-1) + \cost(t_i, g)
    \end{aligned}
    \right. \label{eq:ERP:main}
 \end{align}
\end{subequations}

\subsubsection{Move-Split-Merge}
Move-Split-Merge (MSM, see Equations~\ref{eq:MSM:corner} to \ref{eq:MSM:main}) is a metric developed to overcome shortcomings
in other elastic distances~\citep{stefanMoveSplitMergeMetricTime2013}.
Compared to ERP, it is robust to translation%
\footnote{In ERP, the gap cost is given by $\cost(s_i, g)$.
If $S$ is translated, the gap cost also changes while the canonical alignment cost remains unchanged,
making ERP translation-sensitive.}.
MSM uses its own cost function $C_c$ to compute the cost of the alternate alignments (Equation~\ref{eq:MSMCost}).
It takes as arguments the new point ($\R{np}$) of the alternate alignment,
and the two previously considered points ($x$ and $y$) of each series.
MSM is parameterized by a penalty $c$ involved in the computation of $C_c$.
The authors showed that MSM is competitive against DTW, CDTW and ERP for NN classification.

\begin{equation}\label{eq:MSMCost}
 C_c(\R{np}, x, y) = \left\{
 \begin{aligned}
     & c \quad\text{If $x\leq{}\R{np}\leq{}y$ or $x\geq{}\R{np}\geq{}y$}\\
     & c+ \min\left\{
        \begin{aligned}
        & \abs{\R{np}-x} \\
        & \abs{\R{np}-y} 
        \end{aligned}
        \right. \quad\text{otherwise}
 \end{aligned}
 \right.
\end{equation}

\begin{subequations}\label{eq:MSM}
 \begin{align}
    M_{\MSM}(0,0) &= 0       \label{eq:MSM:corner}\\
    M_{\MSM}(i,0) &= +\infty \label{eq:MSM:vborder}\\
    M_{\MSM}(0,j) &= +\infty \label{eq:MSM:hborder}\\
    M_{\MSM}(i,j) &= \min\left\{
    \begin{aligned}
        &M_{\MSM}(i-1, j-1) + \abs{s_i - t_j} \\
        &M_{\MSM}(i-1, j) + C(s_i, s_{i-1}, t_j) \\
        &M_{\MSM}(i, j-1) + C(t_j, s_i, t_{j-1})
        \end{aligned}
    \right. \label{eq:MSM:main}
 \end{align}
\end{subequations}

\subsubsection{Time Warp Edit Distance}
The Time Warp Edit distance (TWE, see Equations~\ref{eq:TWE:corner} to \ref{eq:TWE:main})
was designed to take time\-stamps, i.e. when a value is recorded,
into account~\citep{marteauTimeWarpEdit2009}.
This matters for series with non-uniform sampling rates.
The $i$-th timestamp of a series $S$ is denoted by $\tau_{S,i}$.
Our current implementation does not use timestamps, 
assuming a constant sampling rate, and we always have $\tau_{s,i} = i$.
TWE defines its own cost functions (Equations~\ref{eq:TWECost}) with two parameters.
The first one, $\nu$, is a ``stiffness'' parameter weighting the timestamp contribution
($\nu=0$ is similar to DTW).
The second one, $\lambda$, is a constant penalty added to the cost of the alternate alignments
(``delete'' in TWE terminology --- deleteA for the lines, deleteB for the columns).
The cost of the alternate case is the cost between the two current points,
plus their timestamp difference, plus the $\lambda$ penalty.
The canonical alignment (``match'') cost is the sum of the cost between the two current
and the two previous points,
plus a weighted contribution of their respective timestamps difference.

\begin{equation} \label{eq:TWECost}
    \begin{aligned}
        \text{match:} &\, \gamma_M = \cost(s_i, t_j) + \cost(s_{i-1}, t_{j-1})
        + \nu(\abs{\tau_{s,i} - \tau_{t,j}} + \abs{\tau_{s, i-1} - \tau_{t, j-1}} ) \\
        \text{deleteA:} &\,  \gamma_A = \cost(s_i, s_{i-1}) + \nu\abs{\tau_{s,i} - \tau_{s, i-1}} + \lambda \\
        \text{deleteB:} &\,  \gamma_B = \cost(t_j, t_{j-1}) + \nu\abs{\tau_{t,j} - \tau_{t, j-1}} + \lambda
    \end{aligned}
\end{equation}

\begin{subequations}\label{eq:TWE}
 \begin{align}
    M_{\TWE}(0,0) &= 0       \label{eq:TWE:corner}\\
    M_{\TWE}(i,0) &= +\infty \label{eq:TWE:vborder}\\
    M_{\TWE}(0,j) &= +\infty \label{eq:TWE:hborder}\\
    M_{\TWE}(i,j) &= \min\left\{
    \begin{aligned}
        &M_{\TWE}(i-1, j-1) + \gamma_M \\
        &M_{\TWE}(i-1, j) + \gamma_A \\
        &M_{\TWE}(i, j-1) + \gamma_B
    \end{aligned}
    \right. \label{eq:TWE:main}
 \end{align}
\end{subequations}

\subsection{\label{sec:Common}Algorithms for the Common Structure}
All the distances presented in Section~\ref{sec:bg:ED} share the structure
captured by Equations~\ref{eq:Common:corner} to~\ref{eq:Common:main}.
These equations are implemented by Algorithm~\ref{alg:Common}, with a $O(L)$ space complexity.
Indeed, a cell $(i,j)$ only depends on the previous row (at $(i-1,j-1)$ or $(i-1,j)$),
or on its left neighbor in the current row (at $(i,j-1)$).
It follows that a row by row implementation only requires two rows of the matrix at any time,
achieving linear space complexity.
Using the shortest series along the columns minimizes the required row length,
further reducing the memory footprint.

\begin{algorithm2e}
    \small
    \SetAlgoLined
    \LinesNumbered
    \DontPrintSemicolon
    \SetKwData{c}{c}
    \SetKwFunction{cost}{cost}
    \KwIn{the time series $S$ and $T$}
    \KwResult{Cost $D(S,T)$}
    \co $\leftarrow$ shortest series between $S$ and $T$\;
    \li $\leftarrow$ longest series between $S$ and $T$\;
    (\prev, \curr) $\leftarrow$ arrays of length $l_{\co}+1$\;
    $\curr{0} \leftarrow 0 $\; \label{alg:Common:init0}
    $\curr{1 --- L} \leftarrow \R{InitHBorder}$\; \label{alg:Common:hborder}
    \For{$i \leftarrow 1$ \KwTo $L_{\li}$}{
      \swap{\prev, \curr}\; \label{alg:Common:swap}
      \curr{0} $\leftarrow \R{InitVBorder}$\; \label{alg:Common:vborder}
      \For{$j \leftarrow 1$ \KwTo $L_{\co}$}{\label{alg:Common:iloop}
        $\curr{j} \leftarrow \min\left\{\begin{aligned}
        &\prev{j-1} &&+ \R{Canonical}  \\
        &\prev{j}   &&+ \R{AlternateRow}  \\
        &\curr{j-1} &&+ \R{AlternateColumn}  \\
        \end{aligned}
        \right.$\;
      }
    }
    \Return{\curr{$l_{\co}$}}\;
    \caption{\label{alg:Common}Generic linear space complexity for a distance $D$.}
\end{algorithm2e}

Two arrays represent the current row (\curr) and the previous row (\prev).
The arrays are swapped at each iteration of the outer loop (line~\ref{alg:Common:swap}),
the current row becoming the previous row,
and the array formerly used for the previous row being assigned for use as the new current row.
Initially, the horizontal border is stored in the \curr array (line~\ref{alg:Common:hborder}).
After the first swap, it will be in \prev, acting as the previous row for the first line.
The vertical border is gradually computed for each new row (line~\ref{alg:Common:vborder}).
Finally, the inner loop computes from left to right
the value of the cells of the current row (line~\ref{alg:Common:iloop}).

Algorithm~\ref{alg:CommonWEA} builds upon Algorithm~\ref{alg:Common},
adding support for a warping window $w$ and the usual early abandoning technique:
it monitors the boundary of the current pass through the cost matrix
and abandons when all values on the boundary exceed a cut-off value.
In this case the boundary is the current row.
Thus, after each row is done, the algorithm looks at the minimum value of the row
and abandons if it is above the cut-off.

\begin{algorithm2e}
    \small
    \SetAlgoLined
    \LinesNumbered
    \DontPrintSemicolon
    \SetKwData{c}{c}
    \SetKwFunction{cost}{cost}
    \KwIn{the time series $S$ and $T$, a warping window $w$, a cut-off value $\cutoff$}
    \KwResult{Cost $D(S,T)$}
    \co $\leftarrow$ shortest series between $S$ and $T$\;
    \li $\leftarrow$ longest series between $S$ and $T$\;
    \lIf{$w < L_{\li} - L_{\co}$}{\Return $+\infty$} \label{alg:CommonWEA:checkw}
    (\prev, \curr) $\leftarrow$ arrays of length $L_{\co}+1$ filled with $+\infty$\;\label{alg:CommonWEA:array}
    $\curr{0} \leftarrow 0 $\; \label{alg:CommonWEA:init0}
    $\curr{1 --- w+1} \leftarrow \R{InitHBorder}$\; \label{alg:CommonWEA:hborder}
    \For{$i \leftarrow 1$ \KwTo $L_{\li}$}{
      \swap{\prev, \curr}\;
      $\jStart \leftarrow \max(1, i-w) $\; \label{alg:CommonWEA:jstart}
      $\jStop \leftarrow  \min(i+w, L_{\co})$\; \label{alg:CommonWEA:jstop}
      \curr{$\jStart-1$} $\leftarrow$ \leIf{$\jStart==1$}{$\R{InitVBorder}$}{$+\infty$} \label{alg:CommonWEA:vborder}
      $\R{minv} \leftarrow +\infty$ \;
      \For{$j \leftarrow \jStart$ \KwTo $\jStop$}{ \label{alg:CommonWEA:jloop}
        $v \leftarrow \min\left\{\begin{aligned}
        &\prev{j-1} &&+ \R{Canonical}  \\
        &\prev{j}   &&+ \R{AlternateRow}  \\
        &\curr{j-1} &&+ \R{AlternateColumn}  \\
        \end{aligned}
        \right.$\;
        $\R{minv} \leftarrow \min(\R{minv}, v)$\;
        \curr{j} $\leftarrow v$ \;
      }
    \lIf{$\R{minv}>\cutoff$}{\Return $+\infty$}
    }
    \Return{\curr{$L_{\co}$}}\;
    \caption{\label{alg:CommonWEA} Generic distance with window and early abandoning.}
\end{algorithm2e}

This algorithm also shows how to handle a window $w$.
It first checks whether $w$ permits an alignment or not (line~\ref{alg:CommonWEA:checkw}).
The horizontal border is only initialized up to $w+1$ (line~\ref{alg:CommonWEA:hborder}),
and the inner loop is caped within the window around the diagonal
(from $\jStart$ to $\jStop$, lines~\ref{alg:CommonWEA:jstart},~\ref{alg:CommonWEA:jstop} and~\ref{alg:CommonWEA:jloop}).
The vertical border is only computed while the window covers the first column (line~\ref{alg:CommonWEA:vborder}).
More interestingly, the arrays are now initialized to $+\infty$.
To understand why, let us consider the cell $(2,3)$ for $M_{\CDTW}$ in Figure~\ref{fig:CDTW:ok}.
It depends on $(1,3)$ (AlternateRow case), which is outside the window.
By initializing the arrays to $+\infty$,
we implicitly set this cell, and all the upper right triangle outside of the window, to $+\infty$.
The lower triangle is implicitly set to $+\infty$ line~\ref{alg:CommonWEA:vborder},
after the window stops covering the first column.
This explains why we assign to \curr{$\jStart-1$} and not to \curr{0}.
Only the diagonal edge of the triangle is set to $+\infty$,
which is enough for the next cell's AlternateColumn case to be properly ignored
(e.g. in Figure~\ref{fig:CDTW:ok} the cell $(3,2)$ ignores $(3,1)$).

% --- --- --- --- --- --- --- --- --- --- --- --- --- --- --- --- --- --- --- --- --- --- --- --- --- --- --- --- --- ---
% --- --- --- --- --- --- --- --- --- --- --- --- --- --- --- --- --- --- --- --- --- --- --- --- --- --- --- --- --- ---
\section{\label{sec:relwork}Related Work}
Most previous work on speeding up NN search under elastic distances has focused on 
improving lower bounds for CDTW (and DTW, which is a special case of CDTW)
\citep{sang-wookkimIndexbasedApproachSimilarity2001,
keoghExactIndexingDynamic2005a,mueenExtractingOptimalPerformance2016,tanFastEEFastEnsembles2020,WEBB2021107895}.
Another approach to directly speeding up elastic distances is through approximation.
To our knowledge, this has only been done for CDTW with FastDTW
(see~\cite{salvadorAccurateDynamicTime2007}, and~\cite{wuFastDTWApproximateGenerally2020} for a counterpoint).
Our approach computes exact CDTW (and other distances, unless early abandoned),
which is useful when approximation is not desirable.

Pruning was first developed for CDTW in 2016 with PrunedDTW \citep{silvaSpeedingAllPairwiseDynamic2016}.
It aimed at speeding up all pairwise calculations of exact DTW
when early abandoning is not an option \citep{zhuNovelApproximationDynamic2012}.
It was subsequently extended to incorporate early abandoning \citep{silvaSpeedingSimilaritySearch2018}
using the classical early abandoning technique
of monitoring the boundary of the progress through the cost matrix
until the minimum value on the boundary exceeds the cut-off (similar to Algorithm~\ref{alg:CommonWEA}).
Both PrunedDTW and its early abandoned version tighten the cut-off during the computation,
which we do not cover in this paper.

EAPruned uses the same overarching pruning strategy as PrunedDTW.
Hence, we will first present EAPruned before giving a comparison with PrunedDTW 
(Section~\ref{sec:eapruned:pruneddtw}).
The main differences are that EAPruned tightly integrates pruning and abandoning through a strategy of abandoning
when pruning leaves no path through the matrix open,
and reduces computational effort by organizing the computation in specialized stages,
pruning in a more efficient way.
Finally, EAPruned generalizes to the class of distances encompassed by Equations \ref{eq:Common:corner} to \ref{eq:Common:main}.

% --- --- --- --- --- --- --- --- --- --- --- --- --- --- --- --- --- --- --- --- --- --- --- --- --- --- --- --- --- ---
% --- --- --- --- --- --- --- --- --- --- --- --- --- --- --- --- --- --- --- --- --- --- --- --- --- --- --- --- --- ---
\section{\label{sec:eapruned}The EAPruned technique}
%EAPruned revisits the pruning idea behind PrunedDTW and
%tightly integrates it with early abandoning.
%Moreover, it is carefully designed for computational efficiency,
%being organized in successive stages in which we avoid computing unnecessary dependencies.
%EAPruned also simplifies the PrunedDTW approach (see Section~\ref{sec:eapruned:pruneddtw}).
We first present EAPruned's two strategies for pruning, ``Pruning from the left'' and ``Pruning on the right''. We then present the EAPruned algorithm (Algorithm~\ref{alg:CommonWEAP}) and finally present
the PrunedDTW technique in the light of EAPruned.
Note that when we illustrate EAPruned, we use DTW with a warping window for the sake of simplicity.

EAPruned utilizes a cut-off value, such as usually provided by the similarity search process (Section~\ref{sec:bg-sslb}).
If a cut-off value is not applicable, EAPruned can be used as a prune-only algorithm, just like PrunedDTW,
using an upper bound based on the diagonal of the cost matrix (e.g. the squared Euclidean distance in the case of DTW).
Note that any warping path through the cost matrix can be used as an upper bound:
either it is an optimal path, or an optimal path will have a lower cost.
Such an upper bound allows pruning (some cells of the cost matrix won't be computed)
but not early abandoning (allowing at least the corresponding path to be computed).

\subsection{Pruning From the Left}

\begin{figure}
  \centering
  \subfloat[][\label{fig:DTW:EAMatrix:UB9}$M_{\DTW(S,T)}$ with $\cutoff=9$.]{%
    \includegraphics[trim=0 30 0 0, clip, width=0.46\textwidth]{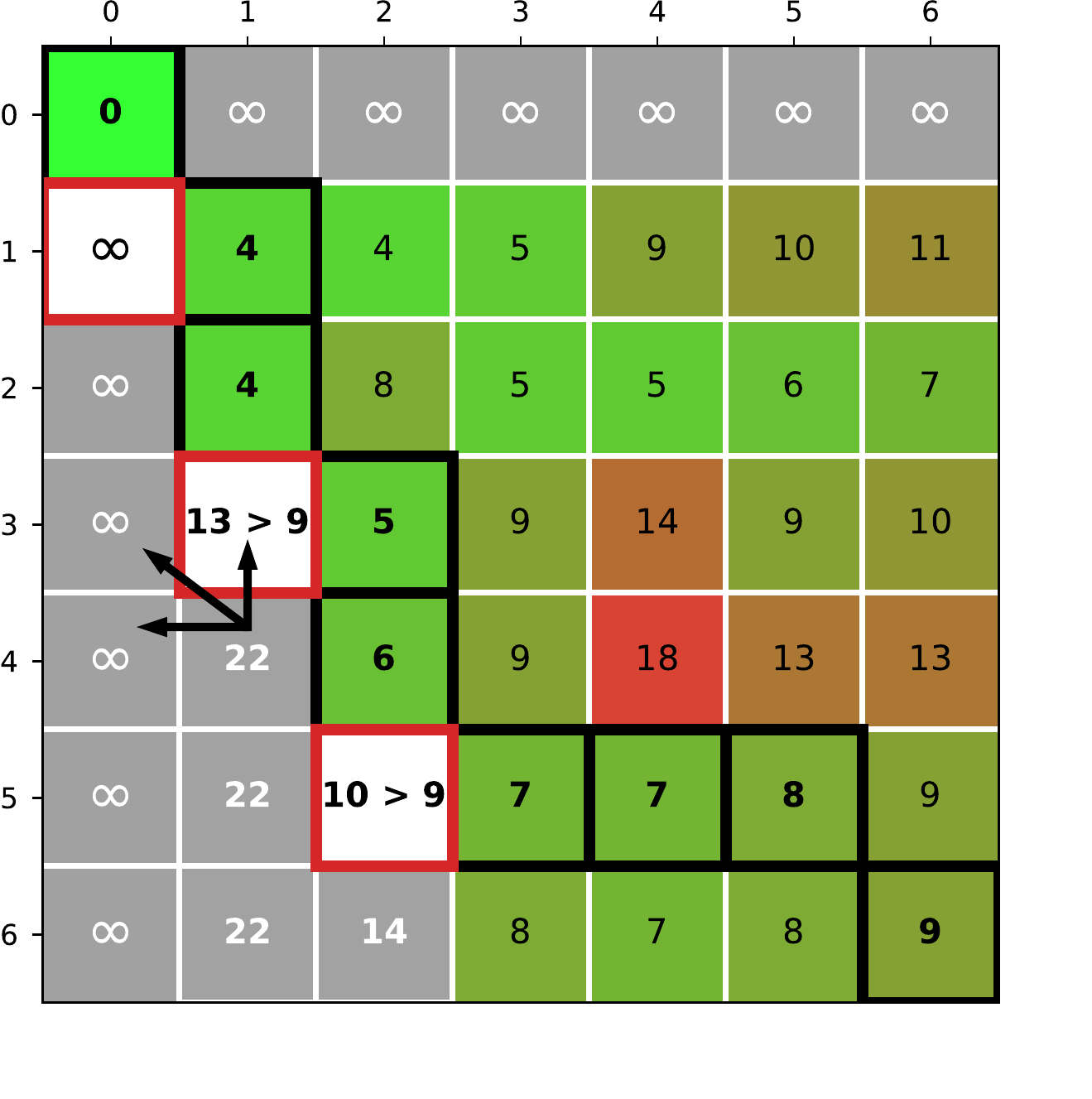}
  }%
  \hspace{\fill}
  \subfloat[][\label{fig:DTW:EAMatrix:UB6}$M_{\DTW(S,T)}$ with $\cutoff=6$.]{%
    \includegraphics[trim=0 30 0 0, clip, width=0.46\textwidth]{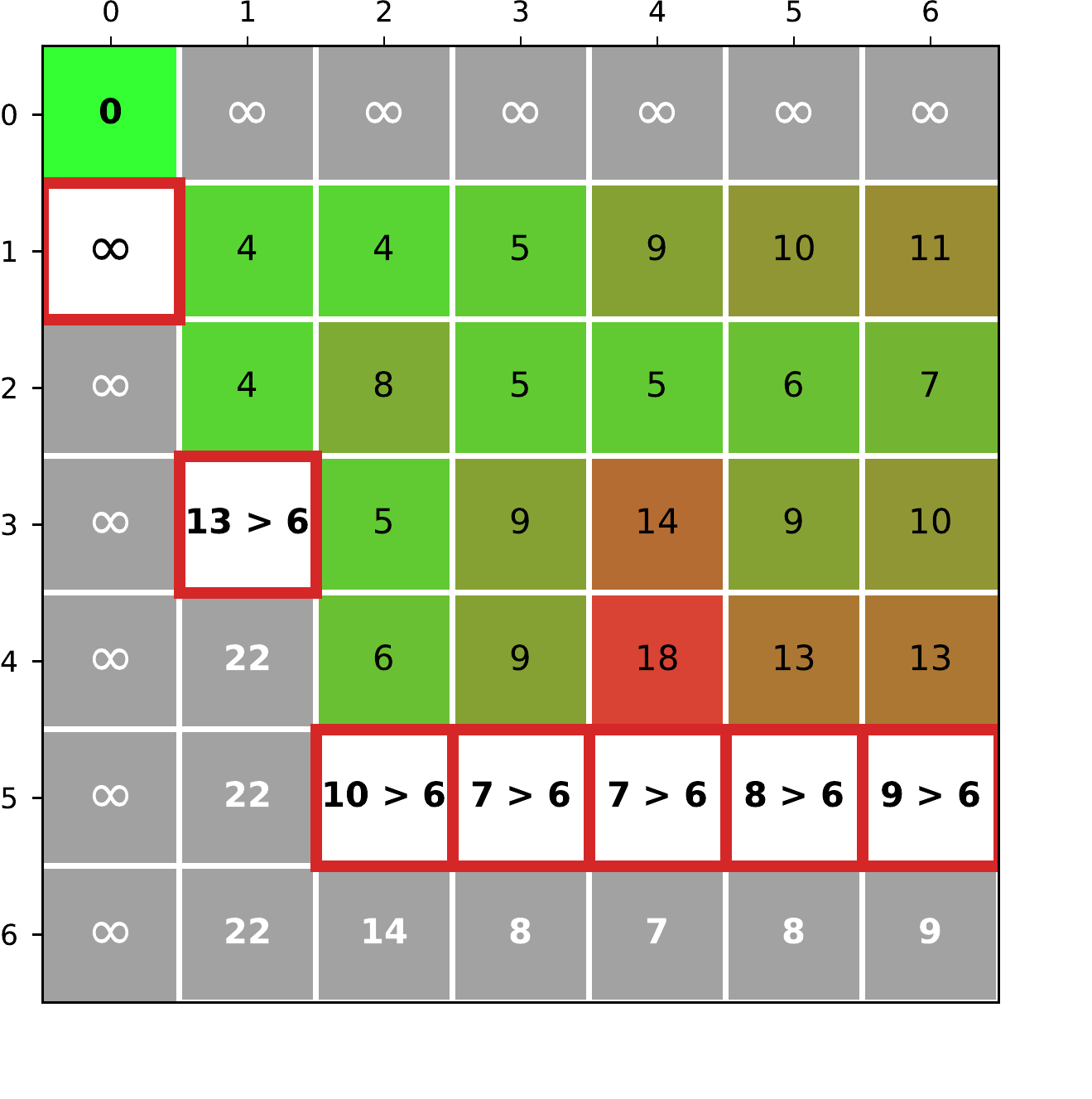}
  }
  \vspace*{-5pt}\caption{\label{fig:DTW:EAMatrix}Illustration of $M_{\DTW(S,T)}$ ``from the left'' pruning
  with two different cut-off values, the second one leading to early abandoning.
  The arrows represent the dependencies of a cell.
  }
\end{figure}

While computing the cost matrix row by row, from left to right,
we look at discarding as many as possible of its leftmost cells.
We define the ``discard zone'' as the ``discarded cells'' topped by a ``discard point'',
i.e. the cells $(i,j)$ such that $\exists_{1\leq k<i}$ such that $(i-k,j)$ is a discard point.
In the $i^{th}$ row, discard points are all cells $(i,l), \ldots , (i,k)$ such that
$\forall_{j\in [l,k]}, M_{D(S,T)}(i,j)>\cutoff$ and $M_{D(S,T)}(i, k+1)\leq\cutoff$,
where $(i,l)$ is the leftmost non discarded cell in the row,
and $\cutoff$ is the cut-off value.
In Figure~\ref{fig:DTW:EAMatrix},
the white cells with red borders are discard points,
and the discard zone is made of the grey cells below them.
If $(i, k+1)$ is out of bound, the computation is early abandoned (Figure~\ref{fig:DTW:EAMatrix:UB6}).

Only the discard points are computed: the discarded cells are pruned (i.e. never computed).
Pruning is enabled by the fact that these cells ultimately depend only on discard points,
which by definition are above the cut-off.
Because a cell can only have a cost greater than (or equal to) its smallest dependency,
cells in the discard zone can only have a cost greater than the cut-off, hence can be ignored.
As an example, take the dependencies of the cell $(4,1)$ in Figure~\ref{fig:DTW:EAMatrix:UB9}.
Starting on the left border, a cell only has a top dependency.
Hence, as soon as a border cell is above the cut-off (i.e. $(0, 1)$),
the remainder of the column can be discarded.
In turn, this creates the same border-like conditions for the next column, starting at the next row.
We only need to check the top dependency for cells $(i>1, 1)$, stopping as soon as possible.
The cell $(3,1)$ is above the cut-off, allowing to discard the remainder of the column, including $(4,1)$.

When proceeding row by row,
pruning from the left is implemented by starting the next row after the discarded columns.
Discarding all the columns in a row, such as in Figure~\ref{fig:DTW:EAMatrix:UB6}, leads to early abandoning.
Note that in this case, the discard points from $(5,3)$ up to $(5,6)$ need to check
both their top and diagonal dependencies.
Most distances have their borders initialized to $\infty$,
starting the discard zone at $(1,0)$ on the left border.
For computed borders, the discard zone starts at the first line $i$ such that $M_{D(S,T)}(i,0)>\cutoff$.
If a window $w$ is used, the discard zone starts at the line $w+1$, unless started earlier
(see Algorithm~\ref{alg:CommonWEAP} line~\ref{alg:CommonWEAP:s1}).

\subsection{Pruning on the Right}

\begin{figure}
  \centering
  \subfloat[][\label{fig:DTW:EAPrunedMatrix:UB9}$M_{\DTW(S,T)}$ with $\cutoff=9$]{%
    \includegraphics[trim=0 30 0 0, clip, width=0.46\textwidth]{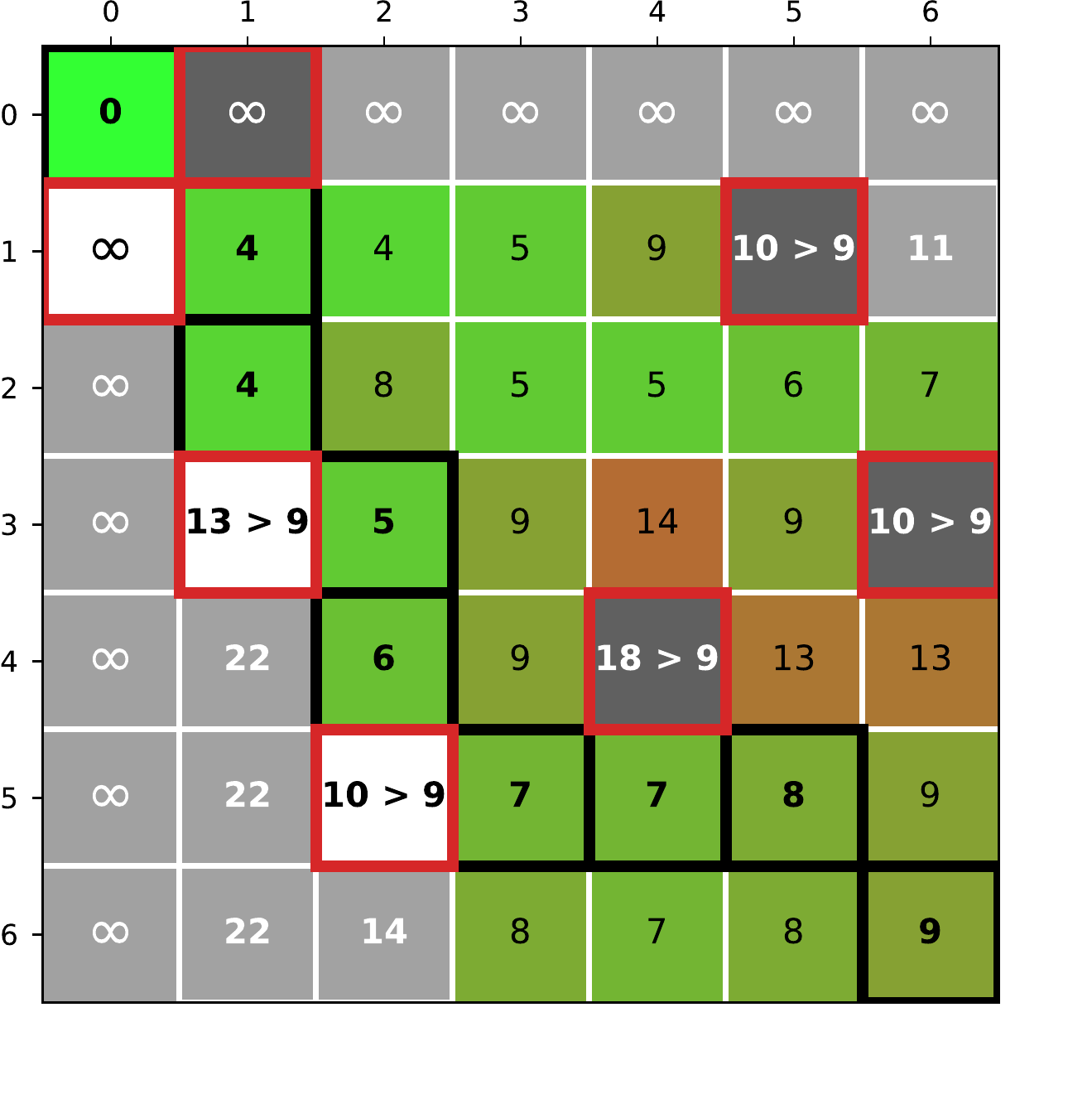}
  }%
  \hspace{\fill}
  \subfloat[][\label{fig:DTW:EAPrunedMatrix:UB6}$M_{\DTW(S,T)}$ with $\cutoff=6$]{%
    \includegraphics[trim=0 30 0 0, clip, width=0.46\textwidth]{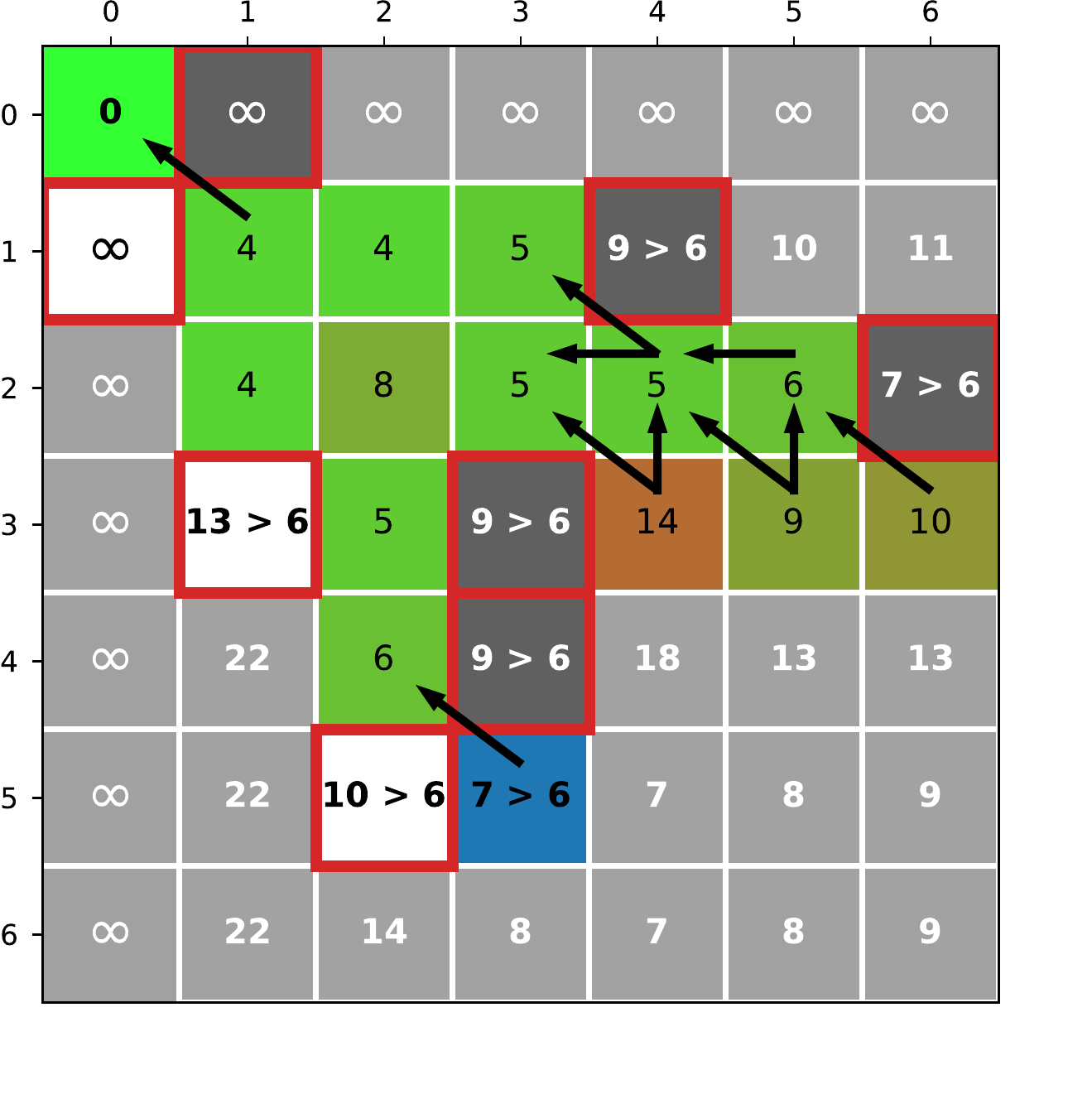}
  }
  \vspace*{-5pt}\caption{\label{fig:DTW:EAPruned}
    Illustration of $M_{\DTW(S,T)}$ ``from the left'' and ``on the right'' pruning
    with two different cut-off values. 
    The blue cell represents the point of early abandoning, white cells represent discard points,
    and dark gray cells represent pruning points.
  }
\end{figure}

In the previous section, we extended a discard zone from the left border to the right,
relying on a row by row evaluation order.
The same is not applicable to the top border, as this would require a column by column evaluation order.
However, we can still find further cells such that all their dependencies are above the cut-off.
Looking back at Figure~\ref{fig:DTW:EAMatrix:UB6}, the cell $(1,4)$ is above the cut-off of 6.
In conjunction with the top border, all the following cells in the row only depend on cells above the cut-off,
hence can be pruned.
Doing so allows us to ``prune on the right''.

In the previous section, we mentioned that a discard point creates the
``same border-like conditions for the next column, starting at the next row'',
perpetuating the pruning process.
Similarly, we have to identify a condition akin to the top border.
This condition is a continuous block of cells above the cut-off value, reaching the end of their row,
i.e. in the $i^{th}$ row $\exists_{1\leq{p_i}\leq{L}},\forall_{p_i\leq{j}\leq{L}}, (i,j)\leq\cutoff$,
with $L$ being the length of the rows.
The start $p_i$ of such a block is called a ``pruning point''.
In Figure~\ref{fig:DTW:EAPrunedMatrix:UB6}, pruning points are represented in dark grey with red borders.
The first one is located on the top border at $(0,1)$, another one is at $(1,4)$.
Pruning points provide information for the next row%
\footnote{Similar to  discard points, instructing the next rows to skip their column.}.
On the next row, as soon as a cell located after the pruning point is above the cut-off,
the remainder of the row can be pruned,
i.e. in the $i^{th}$ row, all the cells $(i,j>c)$ such that $\exists_{c>p_{i-1}}, M_{D(S,T)}(i,c)>\cutoff$.
Note that the cell just under the pruning point is not pruned because of its diagonal dependency
(e.g the cell $(1,1)$ Figure~\ref{fig:DTW:EAPrunedMatrix:UB6}).

Pruning points move back and forth across rows.
To determine their position, a cell $(i,j)$ below the cut-off
will always assume that the next cell is the pruning point of the row.
If $(i,j+1)>\cutoff$, and so on up to $(i, L_{\co})$, then $(i,j+1)$ indeed is a pruning point.
Else, we assume that $(i, j+2)$ is the pruning point, repeating the same logic.
Note that there is only one and only one pruning point,
although it may be out of bounds at $L_{\co}+1$, in which case it does not appear in the figures.
Let us look at some examples in Figure~\ref{fig:DTW:EAPrunedMatrix:UB6}.
The cell $(2,2)>\cutoff$ is not a pruning point because some following cells are below the cut-off.
The cell $(3,3)$ and all its following cells are above the cut-off: it is a pruning point.
Note that the row must be fully evaluated to ensure that $(3,3)$ indeed is the row's pruning point.
It contributes to enabling pruning on the fourth row, starting at $(4,4)$.

Finally, we have to address what happens when both pruning strategies ``collide''
(Figure~\ref{fig:DTW:EAPrunedMatrix:UB6}, in blue).
Without pruning on the right, the cell $(5,3)$ would be a discard point.
Because it is below a pruning point, we know that the rest of the row is above the cut-off,
meaning that the full row is pruned, leading to early abandoning.
If we evaluate a technique by the number of saved computation,
EAPruned saves 16 cells over 36 while the usual early abandoning strategy (Algorithm~\ref{alg:CommonWEA})
only saves the last line, i.e. 6 cells.
Also, if the cell at $(1,1)$ is above the cut-off, EAPruned immediately early abandons
while Algorithm~\ref{alg:CommonWEA} still evaluates the full row.
Finally, even with a cut-off of 9 preventing from early abandoning,
EAPruned still allows to save the computation of 5 cells (Figure~\ref{fig:DTW:EAPrunedMatrix:UB9}).

\subsection{The EAPruned Algorithm}
We now present the EAPruned algorithm (Algorithm~\ref{alg:CommonWEAP}),
which is applicable to any distance matching the common structure described in Section~\ref{sec:bg:ED}.
We present the algorithm with a window and computed borders, covering the most complex case.
EAPruned exploits resulting properties of pruning to further reduce the computational effort.
For example, cells after the pruning point of the previous row
can ignore their top and diagonal dependencies as they are known to be above the cut-off.
To do so, we split the computation of a row in several stages:
\begin{enumerate}
    \item Compute the value of the left border (computed or outside the window).\\
    A computed border may require the top dependency.
    \item Compute discard points until a non-discard point or the pruning point.\\
    Depends on the top and diagonal cells.
    \item Compute non-discard point until the pruning point\\
    Depends on the top, diagonal and left cells.
    \item Deal with the cell at the pruning point\\
    Depends on the diagonal (always) and left (unless was a discard point) cells.
    \item Compute the cells after the pruning point\\
    Depends on the left cell.
\end{enumerate}

\begin{algorithm2e}
    \small
    \SetAlgoLined
    \LinesNumbered
    \DontPrintSemicolon
    \SetKwData{c}{c}
    \SetKwFunction{cost}{cost}
    \KwIn{the time series $S$ and $T$, a warping window $w$, a cut-off point $\cutoff$}
    \KwResult{Cost $D(S,T)$ or $\infty$ if early abandoned}
    \co $\leftarrow$ shortest series between $S$ and $T$\;
    \li $\leftarrow$ longest series between $S$ and $T$\;
    \lIf{$w < L_{\li} - L_{\co}$}{\Return $+\infty$}
    (\prev, \curr) $\leftarrow$ arrays of length $L_{\co}+1$ filled with $+\infty$\;
    $\curr{0} \leftarrow 0$\;
    $\curr{1 --- w+1} \leftarrow \R{InitHBorder}$\;
    $\nextstart \leftarrow 1$ \;
    $\pruningpoint \leftarrow 1$ \;
       \For{$i \leftarrow 1$ \KwTo $L_{\li}$}{
      \swap{\prev, \curr}\;
      $\jStart \leftarrow \max(i-w, \nextstart)$ \;  \label{alg:CommonWEAP:jstart}
      $\jStop \leftarrow  \min(i+w, L_{\co})$ \;
      $j\leftarrow\jStart$\;
      \tcc{Stage 1: init the vertical border} \label{alg:CommonWEAP:s1}
      \curr{$\jStart-1$} $\leftarrow$ \leIf{$\jStart==1$}{$\R{InitVBorder}$}{$+\infty$}
      \tcc{Stage 2: discard points up to, excluding, the pruning point}\label{alg:CommonWEAP:s2}
      \If{$\curr{\jStart-1}>\cutoff$}{\label{alg:CommonWEAP:s2:test}
          \For{$j$ \KwTo $\pruningpoint-1$ \KwWhile $j = \nextstart$}{
              $\curr{j}\leftarrow \min\left\{\begin{aligned}
              &\prev{j-1} && + \R{Canonical} \\
              &\prev{j}   && + \R{AlternateRow}
              \end{aligned}\right.$\;
              \leIf{$\curr{j}\leq\cutoff$}{$\nextpruningpoint\leftarrow j+1$}{$\nextstart\R{++}$}
              \label{alg:CommonWEAP:npp}
          }
      }
      \tcc{Stage 3: continue up to, excluding, the pruning point}\label{alg:CommonWEAP:s3}
      \For{$j$ \KwTo $\pruningpoint-1$}{
        $\curr{j}\leftarrow \min\left\{\begin{aligned}
        &\prev{j-1} &&+ \R{Canonical}  \\
        &\prev{j}   &&+ \R{AlternateRow}  \\
        &\curr{j-1} &&+ \R{AlternateColumn}  \\
        \end{aligned}\right.$\;
        \lIf{$\curr{j}\leq\cutoff$}{$\nextpruningpoint\leftarrow j+1$}
      } 
      \tcc{Stage 4: at the pruning point}\label{alg:CommonWEAP:s4}
      \If{$j\leq{}\jStop$}{
        \If{$j=\nextstart$}{\label{alg:CommonWEAP:s4:dp}
            $\curr{j} \leftarrow \prev{j-1}+\R{Canonical}$\;
            \leIf{$\curr{j}\leq{}\cutoff$}{$\nextpruningpoint \leftarrow j+1$}{\Return{} $+\infty$}
        }
        \Else{\label{alg:CommonWEAP:s4:ndp}
            $\curr{j} \leftarrow \min\left\{\begin{aligned}
            &\prev{j-1} &&+ \R{Canonical}  \\
            &\curr{j-1} &&+ \R{AlternateColumn}  \\
            \end{aligned}\right.$\;
            \lIf{$\curr{j}\leq{}\cutoff$}{$\nextpruningpoint \leftarrow j+1$}
        }
        $j\R{++}$
      }
      \lElseIf{$j=\nextstart$}{\Return{} $+\infty$}\label{alg:CommonWEAP:s4:oob}
      \tcc{Stage 5: after the pruning point}\label{alg:CommonWEAP:s5}
      \For{$j$ \KwTo $\jStop$ \KwWhile $j=\nextpruningpoint$}{
        $\curr{j}\leftarrow \curr{j-1} + \R{AlternateColumn}$ \;
        \lIf{$\curr{j}\leq\cutoff$}{$\nextpruningpoint\leftarrow j+1$}
      }
      $\pruningpoint \leftarrow \nextpruningpoint$\;\label{alg:CommonWEAP:pp}
    } 
    \Return{\curr{$L_{\co}$}}\;
    \caption{\label{alg:CommonWEAP}Generic EAPruned Algorithm.}
\end{algorithm2e}

In Algorithm~\ref{alg:CommonWEAP}, discard points are represented by the \nextstart variable.
The pruning point from the previous row, used to prune in the current row,
is represented by the \pruningpoint variable.
Finally, the \nextpruningpoint variable holds the pruning point being currently computed
(it is assigned to \pruningpoint after a row is done, line~\ref{alg:CommonWEAP:pp}).

Beginning a new row, we first determine the index of the first and last columns.
Then, the first stage updates the left border,
computing its value or applying the window (line~\ref{alg:CommonWEAP:s1}).
The second stage (line~\ref{alg:CommonWEAP:s2}) computes discard points,
which require the row to be bordered on the left by a value above the cut-off
(tested line~\ref{alg:CommonWEAP:s2:test}).
The condition in the for loop ensures that all discard points form a continuous block.
As soon as a value below the cut-off is found,
we jump to the second stage as we cannot have any more discard points.
As explained in the previous section,
\nextpruningpoint is set (line~\ref{alg:CommonWEAP:npp}) to the next column index.
Note that \nextstart can only be updated in the second stages
while \nextpruningpoint must be checked for update after every cost computation.

The third stage (line~\ref{alg:CommonWEAP:s3}) computes
the cost taking all dependencies into account until we reach \pruningpoint.
If \pruningpoint is out of bounds,
the third stage completes the row and the following stages are skipped over.
We enter the fourth stage (line~\ref{alg:CommonWEAP:s4}).
If the row is not done yet, we check if the left cell is a discard point or not.
If it is (line~\ref{alg:CommonWEAP:s4:dp}),
then we only need to check the diagonal dependency,
early abandoning if the resulting cost is above the cut-off.
It not (line~\ref{alg:CommonWEAP:s4:ndp}), both the left and diagonal dependencies are checked.
Finally, we early abandon if the discard points reached the end of the row
(line~\ref{alg:CommonWEAP:s4:oob}).

At the fifth and final stage (line~\ref{alg:CommonWEAP:s5}),
only the left dependency is checked as,
both the diagonal and top dependencies are known to be above the cut-off.
The loop stops as soon as we find a cost above the cut-off, pruning the rest of the row.

\subsubsection{On the Complexity of EAPruned}
Our experiments show that EAPruned achieves significant speed up in several similarity search tasks
(see Section~\ref{sec:experiments}).
As it only allocates two buffers of length $L$, its space complexity is in $O(L)$.
The time complexity depends on the cut-off;
at worst, it remains in $O(L^2)$, at best, it can be as low as $O(1)$
(e.g. in the case of DTW which does not initialize its buffers) or $O(L)$ (with buffer initialization).
This is due to the unpredictable nature of early abandoning under a cut-off.
If the provided cut-off never allows to prune and early abandon (i.e. it is too high),
the full cost matrix will be computed, resulting in a quadratic complexity.
In this case, it is a ``worse'' quadratic complexity than the one from Base due to EAPruned's additional overheads.

Hence, time complexity under early abandoning is a moving target,
sitting between the best and worst case scenarios.
When performing a NN classification, the speed up will depends on the order of evaluation.
If the first candidate is the actual nearest neighbor of a query,
the average complexity will be close to $O(1)$ (or $O(L)$ with buffer initialization)
for all following distance computations.
On the other hand, if the candidates are ordered from the furthest to the nearest of a given query,
the computation will never be early abandoned, and probably not pruned much.
The average complexity will then be closer to $O(L^2)$.

\subsection{\label{sec:eapruned:pruneddtw}Comparison with The PrunedDTW algorithm}

PrunedDTW was not designed with early abandoning in mind,
and hence never considered the case of a $\cutoff$ preventing any alignment.
Furthermore, its extension with early abandoning \citep{silvaSpeedingSimilaritySearch2018}
did not consider pruning in that case either,
instead early abandoning the classical manner (as done in Algorithm~\ref{alg:CommonWEA}).
Algorithm~\ref{alg:PrunedDTW} presents the early abandoned version of PrunedDTW.
This pseudo code is based on the implementation rather
than the pseudo code from \cite{silvaSpeedingSimilaritySearch2018}
that differs from the implementation and appears to be incomplete.
Note that Algorithm~\ref{alg:PrunedDTW} uses $\cutoff'$, a tightened cut-off value based
on the original $\cutoff$ and on an array stored during the lower bounding process representing a lower bound on the remaining alignment for each row of the matrix (``cumLB'').
$\cutoff'$ is tightened several times during computation (lines~\ref{alg:prunedDTW:ub1} and~\ref{alg:prunedDTW:ub2}).
We refer the reader to \cite{silvaSpeedingSimilaritySearch2018} for an explanation of the technique.

\begin{algorithm2e}
    \small
    \SetAlgoLined
    \LinesNumbered
    \DontPrintSemicolon
    \SetKwData{c}{c}
    \SetKwData{F}{False}
    \SetKwData{T}{True}
    \SetKwData{sc}{next\_start}
    \SetKwData{lp}{last\_pruning}
    \SetKwData{ec}{pruning\_point}
    \SetKwData{ecnext}{next\_pruning\_point}
    \SetKwData{foundSC}{foundSC}
    \SetKwData{prunedEC}{prunedEC}
    \SetKwData{minc}{min\_cost}
    \SetKwData{cumLB}{cumLB}
    \SetKw{break}{break}
    \SetKw{continue}{continue}
    \SetKwFunction{cost}{cost}
    \KwIn{the time series $S$ and $T$ of length $L$, a warping window $w$, a cut-off point $\cutoff$,
        cumulative lower bound values \cumLB
    }
    \KwResult{Cost $\DTW(S,T)$ or $\infty$ if early abandoned}
    $(\sc, \ec) \leftarrow (0,0)$\;
    (\prev, \curr) $\leftarrow$ arrays of length $L$ filled with $+\infty$\;
    $\cutoff' \leftarrow \cutoff - \cumLB[w+1]$\label{alg:prunedDTW:ub1}\tcp*{$\cutoff$ tightening}
    \For{$i \leftarrow 0$ \KwTo $L-1$}{
        $\minc \leftarrow \infty$\;
        $(\foundSC, \prunedEC) \leftarrow (\F, \F)$\;
        $\ecnext \leftarrow i+w+1$\;
        $\jStart \leftarrow \max(0, \sc, i - w)$\label{alg:PrunedDTW:linestart}\;
        \For{$j\leftarrow \jStart$ \KwTo $\min(i+w, L-1)$}{
            \tcc{First cell in the cumulative matrix}
            \If{$i=0 \land j=0$}{
                $\curr[0] \leftarrow \cost(s_i, t_j)$ \;
                $\minc \leftarrow \curr[0]$ \;
                $\foundSC \leftarrow \T$\;
                \continue
            }
            
            \tcc{Compute cost excluding invalid cells}\label{alg:PrunedDTW:cost}
            \leIf{$j=\jStart$}{$y\leftarrow\infty$}{$y\leftarrow\curr[j-1]$}
            \leIf{$i=0 \lor j=i+w \lor j \geq \lp$}{$x\leftarrow\infty$}{$x\leftarrow\prev[j]$}
            \leIf{$i=0 \lor j=0 \lor j>\lp$}{$z\leftarrow\infty$}{$z\leftarrow\prev[j-1]$}
            
            $\curr[j]\leftarrow \cost(s_i, t_j) + \min(x,y,z)$\;
            $\minc \leftarrow \min(\minc, \curr[j])$\label{alg:PrunedDTW:checkmin1}\;
            
            \tcc{Pruning criteria}
            \lIf{$\foundSC = \F \land \curr[j]\leq \cutoff'$}{ $(\sc, \foundSC) \leftarrow (j, \T)$ }
            \uIf{$\curr[j]>\cutoff'$}{
                \If(\label{alg:PrunedDTW:checkpp}){$j>\ec$}{
                    $(\lp,\prunedEC)\leftarrow (j, \T)$\;
                    \break
                }
            }\lElse{ $\ecnext \leftarrow j+1$ }
        }
        \tcc{End of inner for loop - Early abandoning and updates}
        \If{$i+w<L-1$}{
            $\cutoff' \leftarrow \cutoff - \cumLB[i+w+1]$\label{alg:prunedDTW:ub2}\;
            \lIf{$\minc > \cutoff'$}{\Return{$\infty$}}\label{alg:PrunedDTW:checkmin2}
        }
        \swap{\curr, \prev}\;
        \lIf{$\sc>0$}{$\prev[\sc-1]\leftarrow\infty$}
        \lIf{$\prunedEC = \F$}{$\lp \leftarrow i+w+1$}
        $\ec \leftarrow \ecnext$\:
    }
    \tcc{Check if the last row was pruned before returning}
    \lIf{$\prunedEC = \T$}{$\curr[j] \leftarrow \infty$}
    \Return{$\curr[L]$}
\caption{\label{alg:PrunedDTW}Early Abandoned PrunedDTW Algorithm.}
\end{algorithm2e}

EAPruned uses the same overarching strategy as PrunedDTW,
which can be tracked by following some shared variable names (e.g. \nextstart and \pruningpoint).
However, EAPruned has the ability to prune more cells than PrunedDTW.
PrunedDTW in essence misses all the stage 4 pruning from Algorithm~\ref{alg:CommonWEAP}.
It only checks what happens after the pruning point (line~\ref{alg:PrunedDTW:checkpp}),
and not at the pruning point,
instead relying on the minimum value in the line to early abandon
(lines~\ref{alg:PrunedDTW:checkmin1} and~\ref{alg:PrunedDTW:checkmin2}).
More importantly, PrunedDTW only uses the discard points to determine the start of the next line
(line~\ref{alg:PrunedDTW:linestart}).
By using this information earlier (computed at stage 2, used at stage 4),
EAPruned is also able to early abandon earlier.

EAPruned not only prunes more cells,
it also does so in a much more efficient way thanks to its staged approach (Section~\ref{exp:lb}).
PrunedDTW always checks all the dependencies of a cell when computing its cost
(starting at line~\ref{alg:PrunedDTW:cost}).
Doing so requires ``sanitizing'' every access in order to exclude pruned cells.
Hence, it not only tests unnecessary dependencies, it actually spends time to do so.
This has a significant impact in a loop body literally executed billions of times across our benchmark.

EAPruned is also more direct and arguably simpler, getting rid of more than half the temporary variables.
EAPruned was first developed for DTW.
Its clean structure naturally leads to a the generalized version presented in this paper.
To the best of our knowledge no generalisation of PrunedDTW to other distance measures has been proposed.

\section{\label{sec:experiments}Experiments}
We evaluate EAPruned in the context of NN search,
which naturally supports early abandoned distances.
In this context, all the presented implementations of a distance produce the exact same NN search results.
Hence, our experiments are about execution speed (and not, e.g., accuracy).
All our experiments have been run under the same conditions, on a computer equipped with 64GB of RAM
(enough memory to fit all the relevant data) and an AMD Opteron 6338P at 2.4Ghz.
The C++ source code for all the implementations (including PrunedDTW)
is available on github~\citep{ALLCODE}.

Our 3 experiments evaluate ``EAPruned'' (Algorithm~\ref{alg:CommonWEAP})
distance implementations against several others:
``Base'' are the classical double buffered implementations, without early abandoning
(Algorithm~\ref{alg:Common});
``EABase'', are Base implementations with the usual early abandoning technique (Algorithm~\ref{alg:CommonWEA}).
``Prune'' are the same as ``EAPruned'',
but always use their own computed cut-off based on the diagonal of the cost matrix
(only allowing to prune, not to early abandon);
Finally, ``PrunedDTW'' and ``PrunedDTW+EA'' are the reference C++ implementations
from \cite{silvaSpeedingAllPairwiseDynamic2016} and \cite{silvaSpeedingSimilaritySearch2018}.
Our experiments provide the information required by ``PrunedDTW+EA'' to tighten the cut-off
(Algorithm~\ref{alg:PrunedDTW}).

\subsection{\label{exp:nn1}Evaluation Under NN Classification}
\begin{figure}
  \centering
  \subfloat[][All distances]{\label{fig:expdist:all}%
    \includegraphics[trim=20 25 0 0, clip, width=0.31\textwidth]{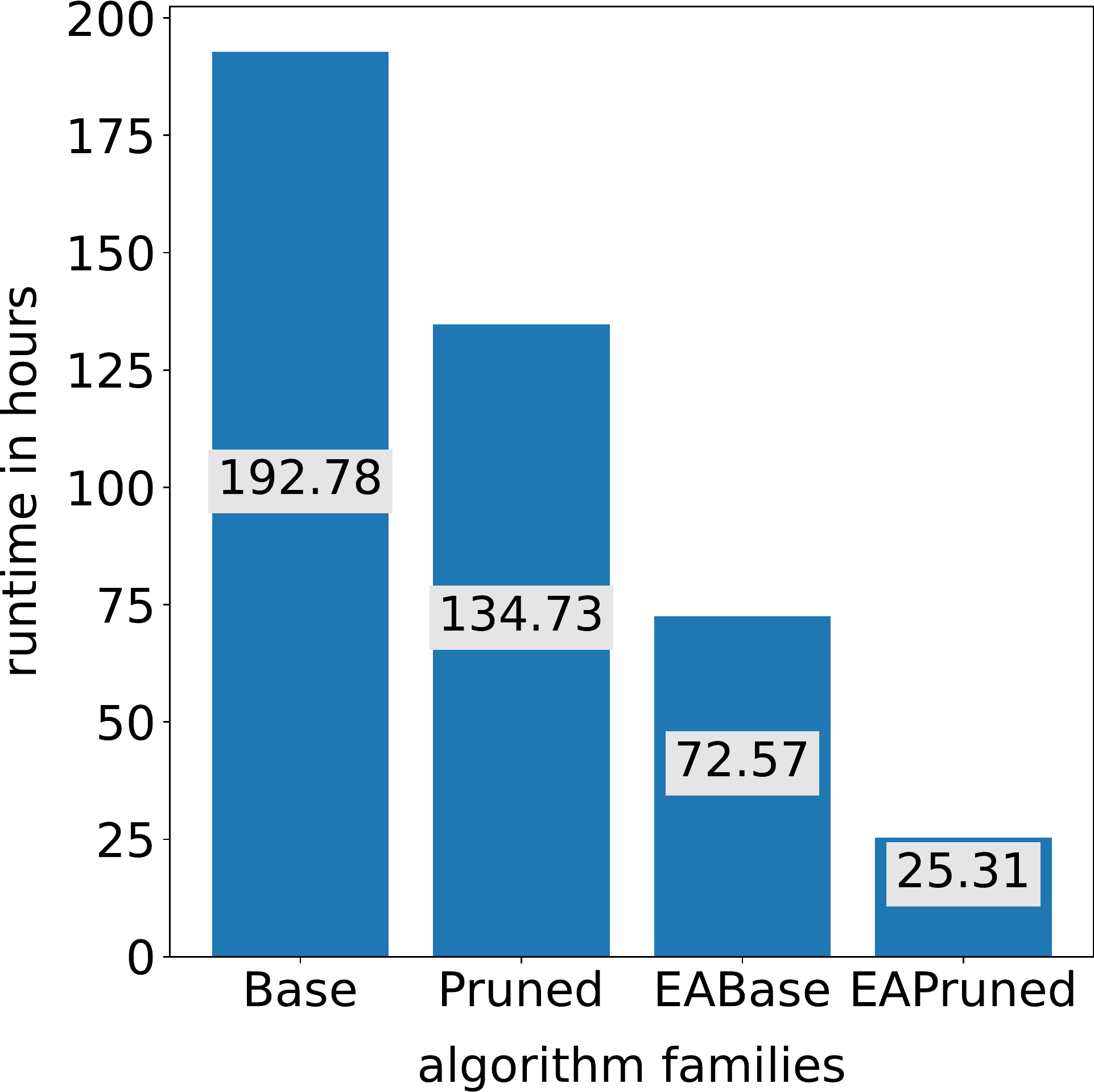}
  }%
  \hspace{\fill}
  \subfloat[][DTW]{\label{fig:expdist:dtw}%
    \includegraphics[trim=20 25 0 0, clip, width=0.31\textwidth]{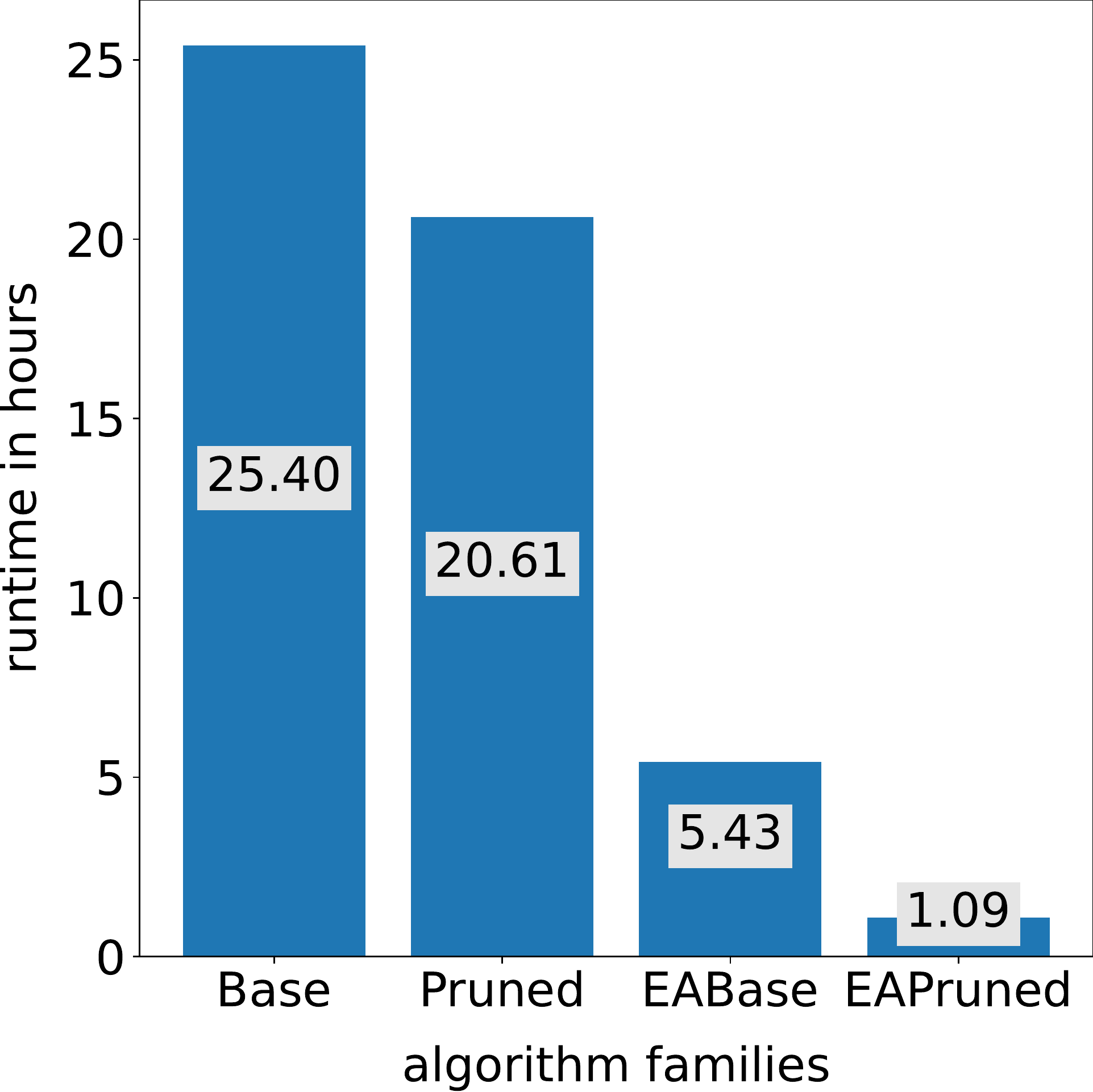}
  }%
  \hspace{\fill}
  \subfloat[][CDTW]{\label{fig:expdist:cdtw}%
    \includegraphics[trim=20 25 0 0, clip, width=0.31\textwidth]{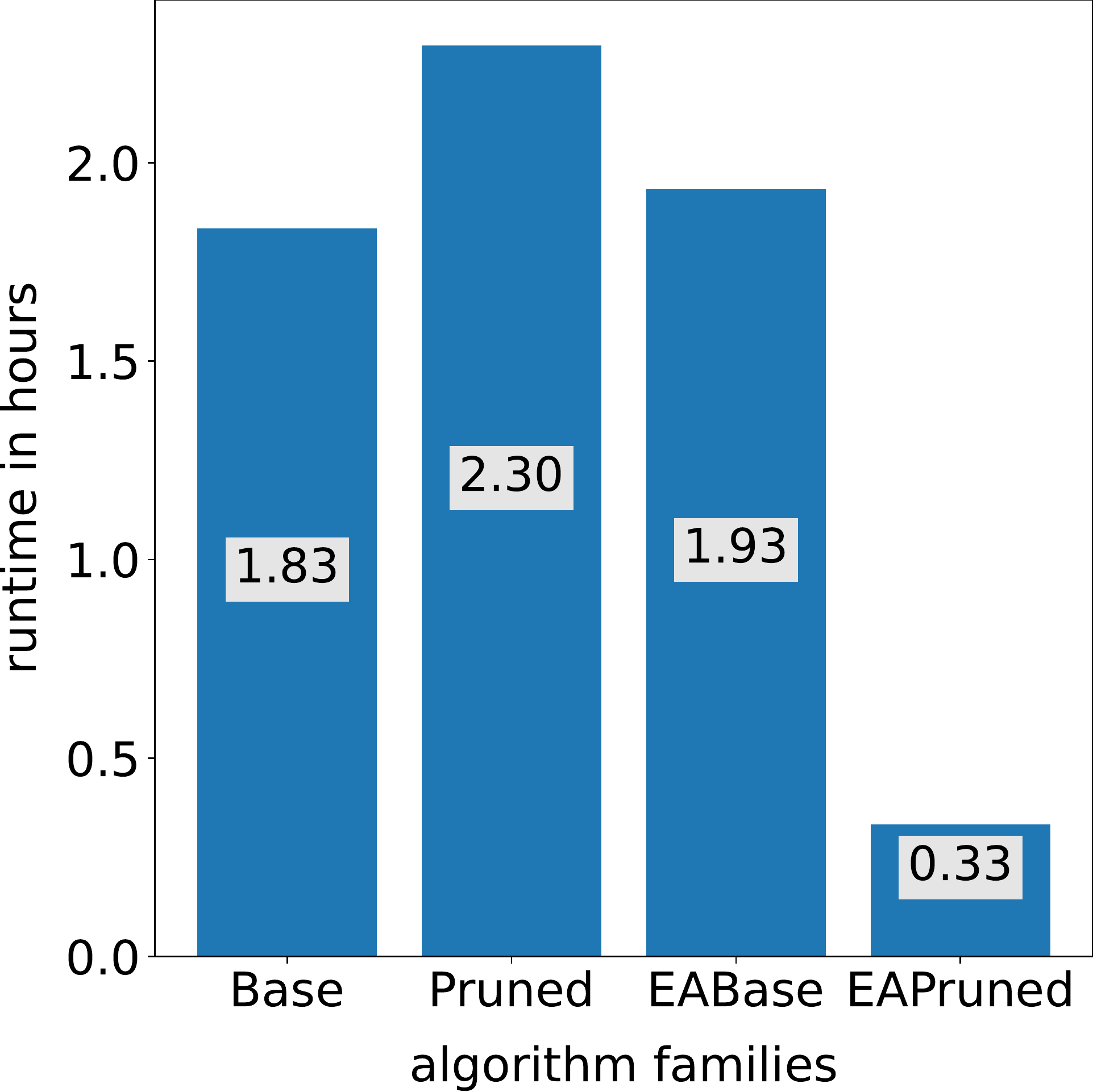}
  }%
  \vspace{0.1\floatsep}
  \subfloat[][WDTW]{\label{fig:expdist:wdtw}%
    \includegraphics[trim=20 25 0 0, clip, width=0.31\textwidth]{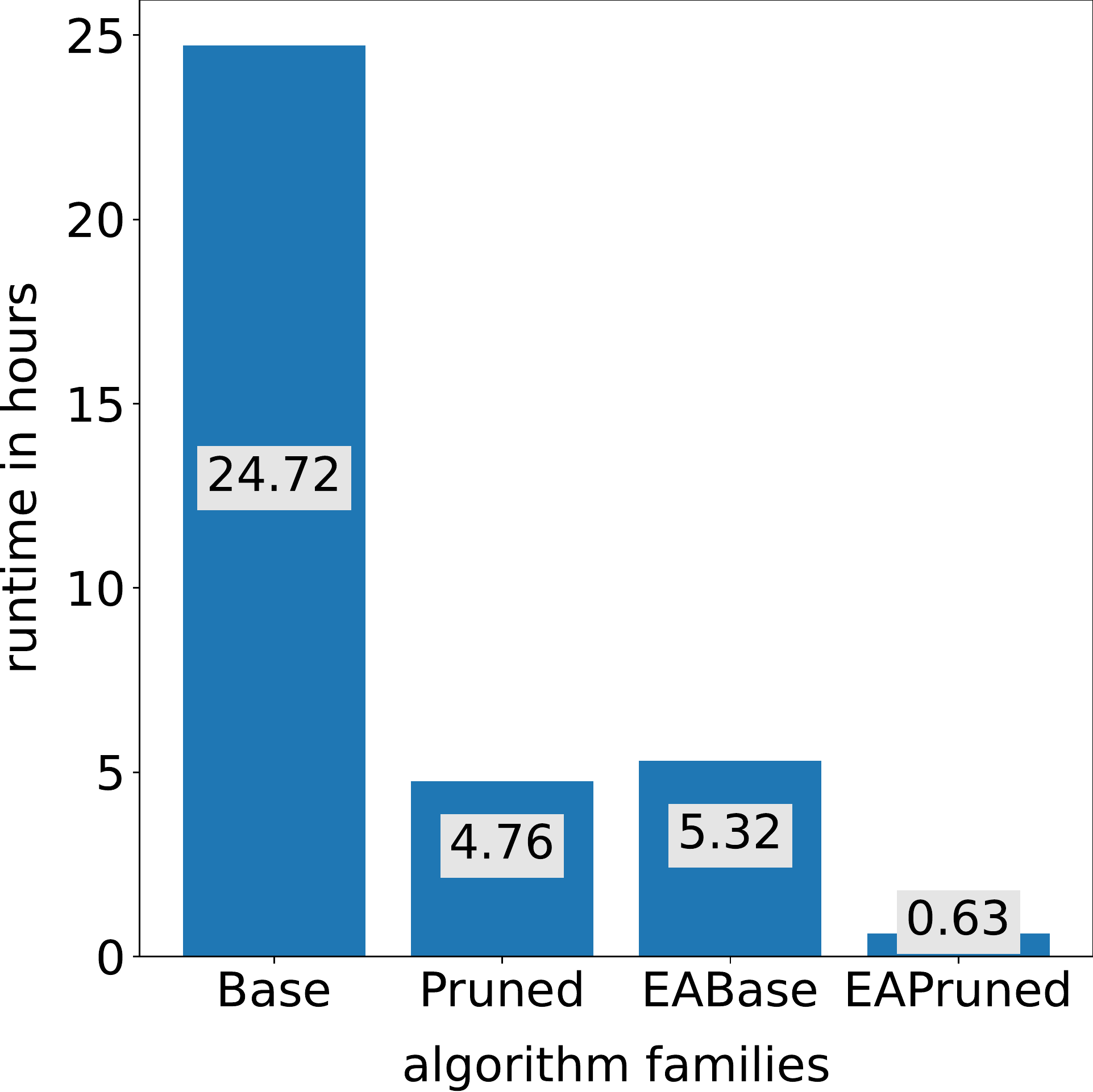}
  }%
  \hspace{\fill}
  \subfloat[][ERP]{\label{fig:expdist:erp}%
    \includegraphics[trim=20 25 0 0, clip, width=0.31\textwidth]{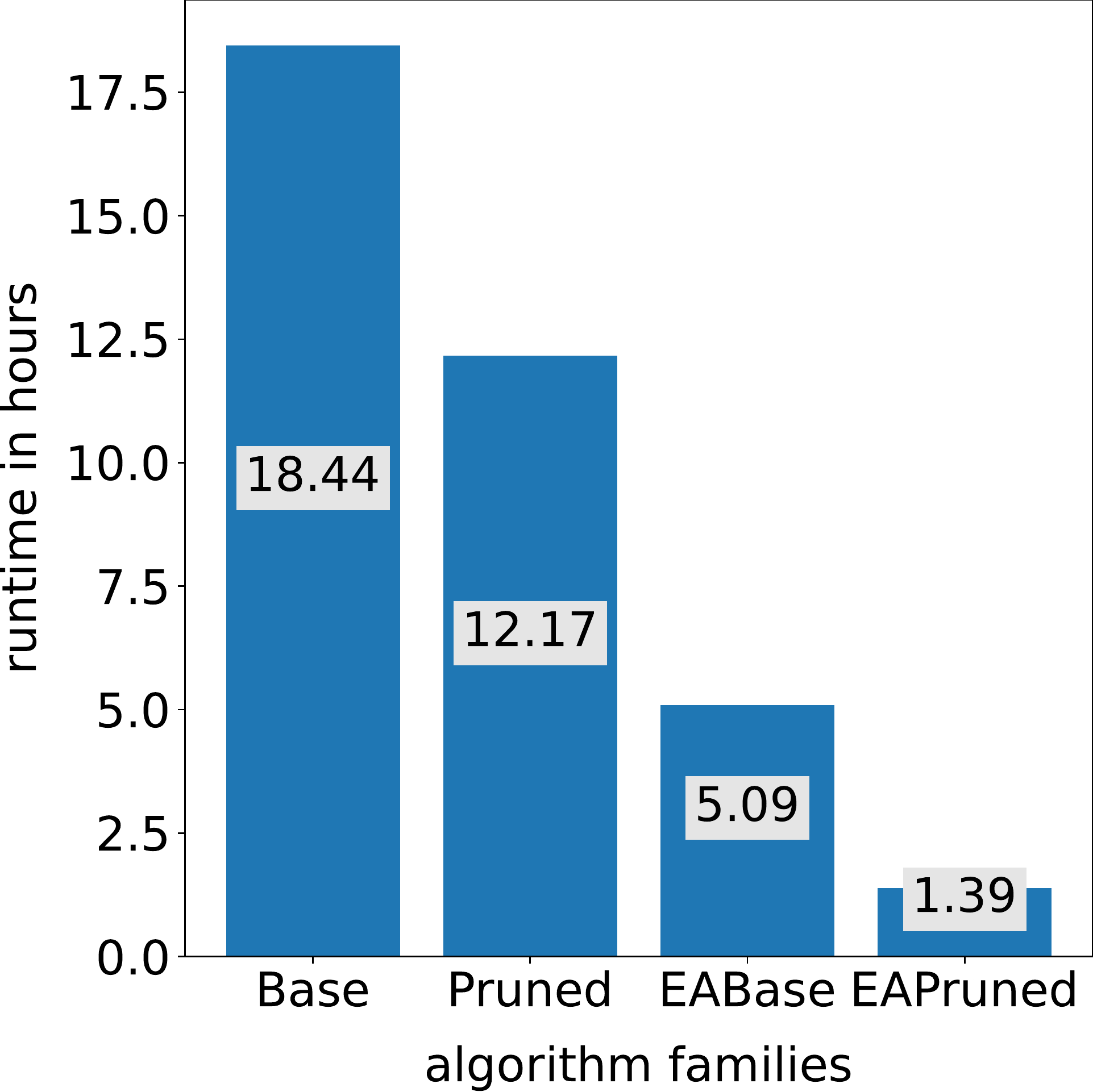}
  }%
  \hspace{\fill}
  \subfloat[][MSM]{\label{fig:expdist:msm}%
    \includegraphics[trim=20 25 0 0, clip, width=0.31\textwidth]{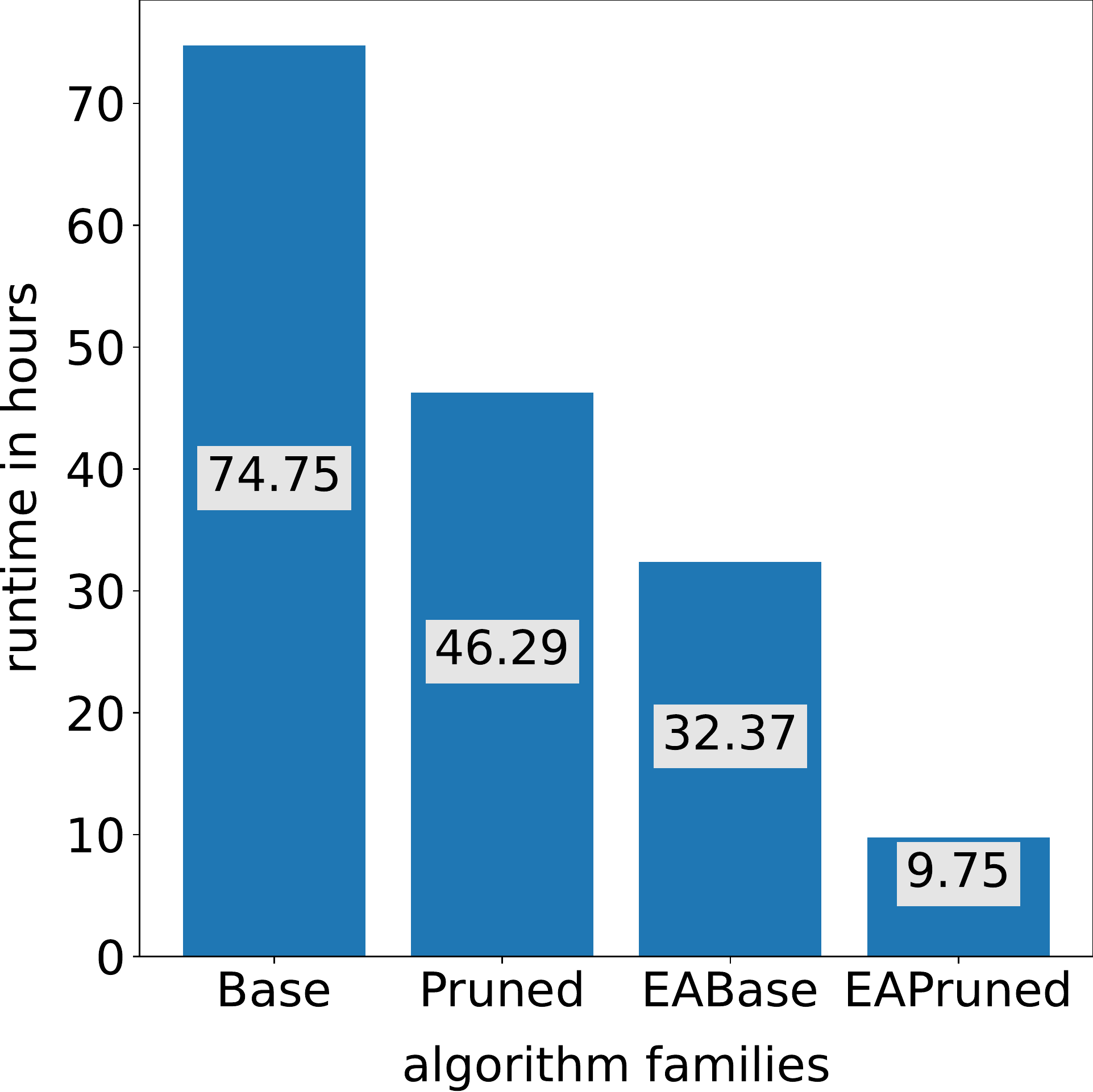}
  }%
  \vspace{0.1\floatsep}
   \subfloat[][TWE]{\label{fig:expdist:twe}%
    \includegraphics[trim=20 25 0 0, clip, width=0.31\textwidth]{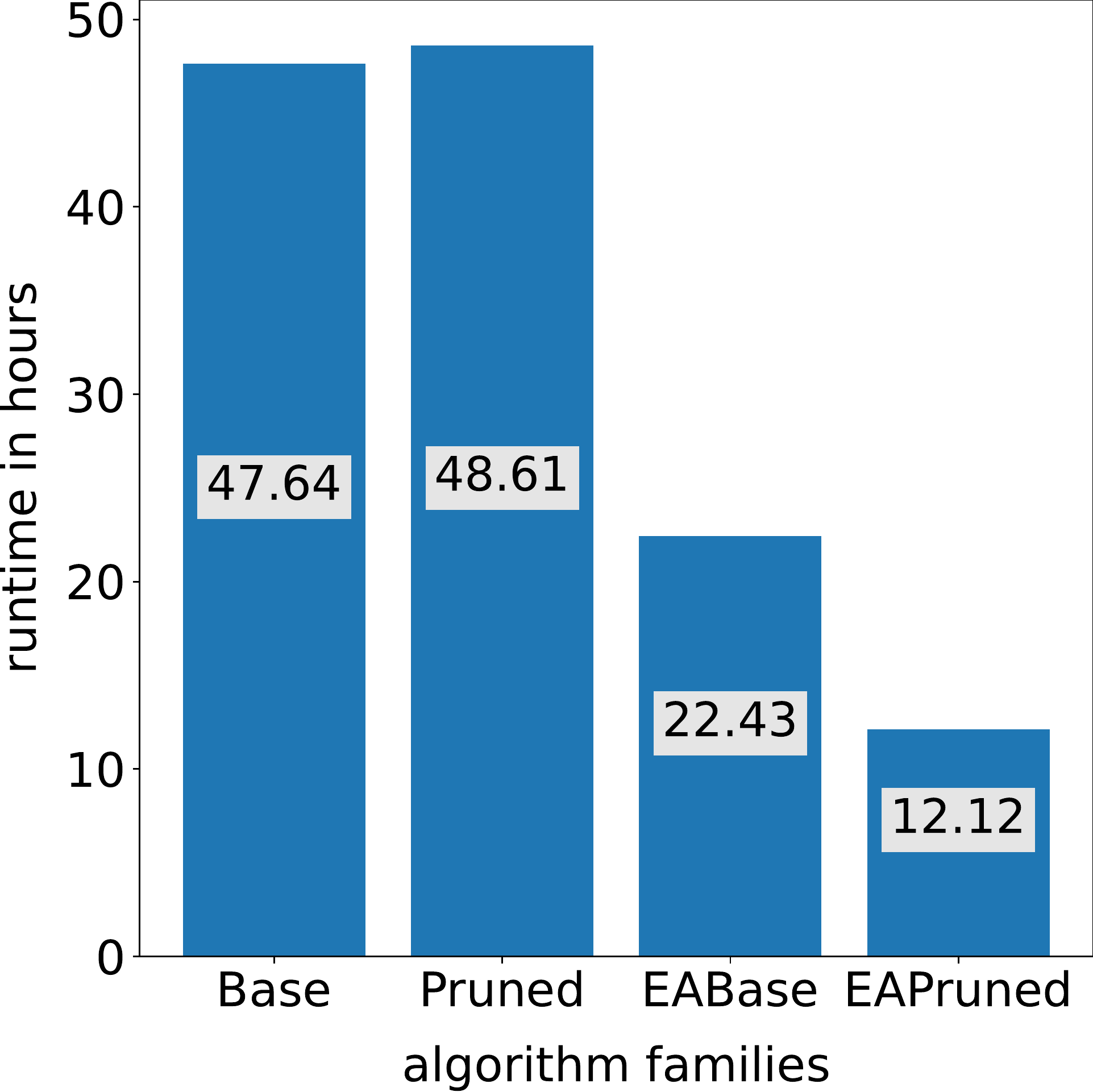}
  }%
  \vspace*{-3pt}\caption{\label{fig:expdist}
  Accumulated timings in hours of NN classification over 85 datasets from the UCR archive,
  using parameters discovered by EE.}
\end{figure}

We compare the timings of the different distance implementations under NN classification.
We use the default train/test splits of the 85 datasets from the UCR Archive \citep{dauUCRTimeSeries2019}.
The Figure~\ref{fig:expdist} presents the results in hours per distance.
We use realistic parameters for the distances, using the one found by EE for each dataset.
This is particularly relevant for distances with a warping window,
as small windows (which is the most common case) greatly impact the runtime,
e.g. CDTW Base is $\approx 12$ times faster than DTW Base
(Figures~\ref{fig:expdist:cdtw} and~\ref{fig:expdist:dtw}).

EAPruned is overall the most efficient implementation (Figure~\ref{fig:expdist:all}),
being $\approx 7.61$ times faster than Base, and $\approx 2.86$ times faster than EABase.
This is confirmed when looking at each individual distances (Figure~\ref{fig:expdist}),
with a speedup ranging from $\approx 3.93$ (TWE) up to $\approx 39.23$ (WDTW) compared to Base,
and from $\approx 1.85$ (TWE) up to $\approx 8.44$ (WDTW) compared to EABase.
Note that only EAPruned achieves a speed up for CDTW compared to the Base version.
Looking at pruning only scenarios (e.g. when all pairwise distance computations are required),
always choosing Prune over Base may not be the best choice.
If Pruned is overall ahead of Base, this is not the case for CDTW and TWE (Figures~\ref{fig:expdist:cdtw} and~\ref{fig:expdist:twe}).

We presented the timings for 85 datasets.
However, the majority of the computation time comes from the slowest datasets.
The 5 slowest datasets ($5.8\%$ of the datasets) make up for $\approx 72.5\%$ of the total Base time
(Table~\ref{tab:slow}, in hours).
Slower datasets have long series, favouring EAPruned,
e.g. StarLightCurves contains 1000 train series and 8236 test series of length 1024.
In this circumstances, EAPruned achieves a greater speedup over Base
($\approx 10.6$) and EABase ($\approx 3.23)$.

EAPruned remains beneficial when looking at the 5 fastest datasets (Table~\ref{tab:fast}, in minutes),
albeit with a smaller speedup ($\approx 2.19$ for Base, $\approx 1.73$ for EABase).
This comes at no surprise as datasets with smaller series 
(ItalyPowerDemand contains 67 train series and 1029 test series of length 24)
offers less pruning opportunities, hence less chances for EAPruned to make up for its overhead.
It nonetheless is the fastest implementation for all the dataset in the archive.

\begin{table}
\centering
\begin{tabular}{|c | c c c c|} 
 \hline
 Dataset & Base & Pruned & EABase & EAPruned \\
 \hline
 NonInvasiveFetalECGThorax2 &16.72 &9.53 &3.81 &0.38 \\
NonInvasiveFetalECGThorax1 &19.23 &7.34 &3.87 &0.36 \\
HandOutlines &19.82 &10.05 &6.89 &1.33 \\
UWaveGestureLibraryAll &21.05 &17.15 &8.44 &3.47 \\
StarLightCurves &63.01 &47.63 &19.58 &7.62 \\
\hline
total &139.83 &91.70 &42.59 &13.16 \\
\hline
\end{tabular}
\caption{The 5 slowest datasets, timings in hours.}
\label{tab:slow}
\end{table}

\begin{table}
\centering
\begin{tabular}{|c | c c c c|} 
 \hline
 Dataset & Base & Pruned & EABase & EAPruned \\
 \hline
 ItalyPowerDemand & 0.029 &0.024 &0.015 &0.012 \\
Coffee & 0.035 &0.029 &0.032 &0.017 \\
SonyAIBORobotSurface1 & 0.051 &0.028 &0.037 &0.016 \\
BirdChicken & 0.060 &0.065 &0.056 &0.030 \\
BeetleFly & 0.060 &0.050 &0.046 &0.032 \\
 \hline
total &0.235 &0.195 &0.186 &0.107 \\
 \hline
\end{tabular}
\caption{The 5 fastet datasets, timings in minutes.}
\label{tab:fast}
\end{table}

\subsection{\label{exp:lb}Evaluation Under NN Classification with Lower Bounds}
\begin{figure}
  \centering
  \includegraphics[trim=0 15 0 33, clip,width=0.999\linewidth]{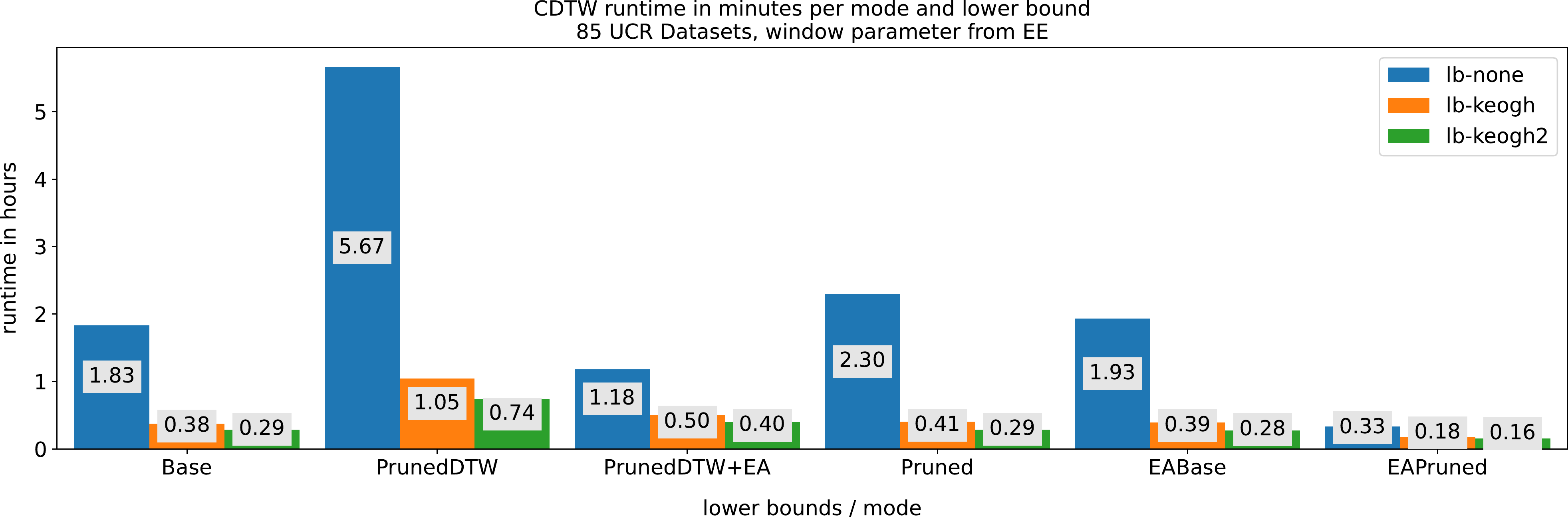}
  \vspace*{-8pt}\caption{\label{fig:explb:CDTW}
  Comparison of NN timings in hours of various CDTW implementations over the UCR Archive,
  under various lower bounds.
  Window parameters obtained from EE.}
\end{figure}

\begin{figure}
  \centering
  \includegraphics[trim=0 15 0 33, clip,width=0.999\linewidth]{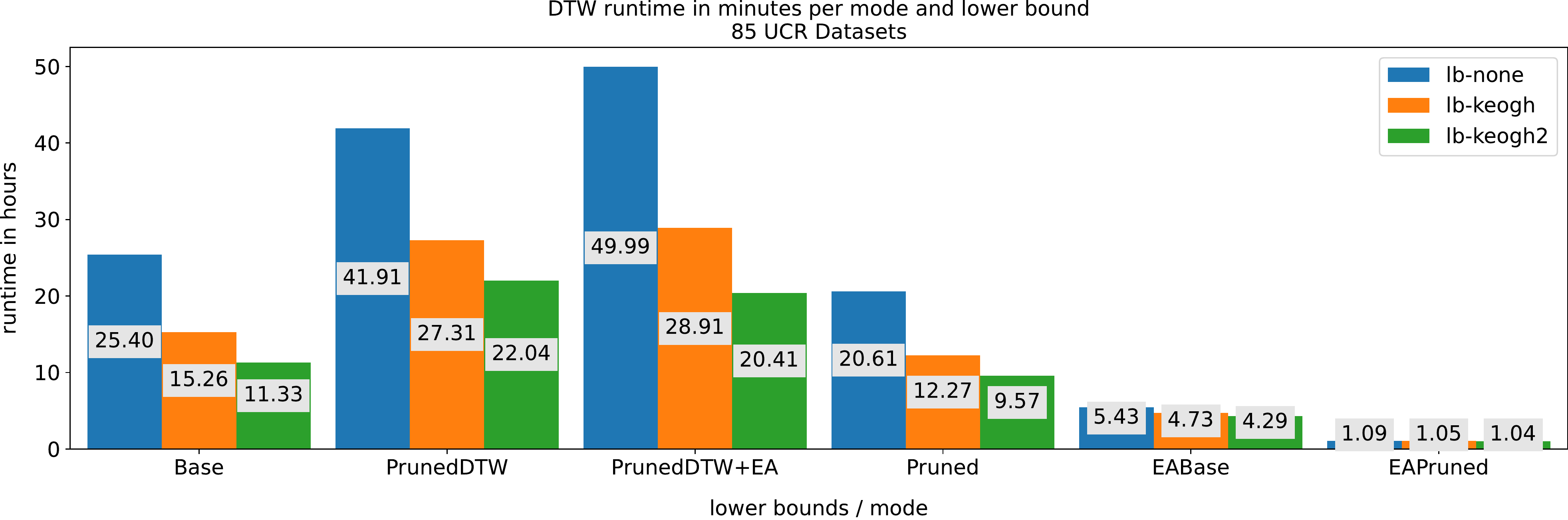}
  \vspace*{-8pt}\caption{\label{fig:explb:DTW}
  Comparison of NN timings in hours of various DTW implementations over the UCR Archive,
  under various lower bounds.
  }
\end{figure}

The previous experiment evaluates NN classification only using the distances.
However, most NN classification scenarios are sped up using lower bounding.
Under these circumstances, is EAPruned still beneficial?
In the following experiment, we focus on DTW and CDTW for which efficient lower bounds exist.
We repeat the previous experiment for CDTW and DTW
(Figures~\ref{fig:explb:CDTW} and~\ref{fig:explb:DTW})
without lower bound (``lb-none''),
the LB-Keogh lower bound (``lb-keogh''),
and cascading two applications of LB-Keogh
(``lb-keogh2'', reversing their arguments as $\R{Keogh}(a,b) \neq \R{Keogh}(b,a)$, 
see~\cite{mueenExtractingOptimalPerformance2016}).
Note that LB-Keogh requires to compute the envelopes of the series.
For a given run, envelopes are only computed once,
in $O(L)$ using Lemire's algorithm~\citep{lemireFasterRetrievalTwopass2009},
keeping this overhead to a minimum.
We also use the occasion to compare EAPruned with ``PrunedDTW'' and ``PrunedDTW+EA''.

In the DTW case, EAPruned is more than 4 times faster than the EABase with lb-keogh2,
one of the fastest configurations known until now.
Lower bounding EAPruned still offers $\approx 10\%$ speedup,
which is interesting as envelopes computed
over a window as wide as the series does not contain much information
(see how the CDTW benefits way more from lower bounding in the Base and EABase cases than DTW).
In the CDTW case, EAPruned without lower bounding is on par with Base and EABase with lower bounds.
Without lower bounds, PrunedDTW+EA comes second, but loses its advantage in other cases.
PrunedDTW is actually slower than the Base version.
It is also slower than our Pruned version even though we do not tighten the cut-off during computation.
Lower bounding EAPruned provides a further $\approx 2$ times speed up.

Our results indicate that lower bounding --- at least with lb-Keogh --- complements EAPruned.

\subsection{\label{exp:ss}Sub-sequence Search}
The two previous experiments present NN classification results, which are an application of NN search.
Another kind of similarity search is sub-sequence search.
Given a query, sub-sequence search locates the closest match (in our case, under DTW)
within another, usually very long, reference series.
The UCR suite \citep{rakthanmanonSearchingMiningTrillions2012} is the first scalable tool for the task,
deploying several optimisations, including a custom early abandoned DTW.
It later evolved into the UCR-USP suite, including ``PrunedDTW+EA''\citep{silvaSpeedingSimilaritySearch2018}.
We replaced the DTW implementation of the UCR suite with our own EAPruned DTW,
and replicated the experiments from \citep{silvaSpeedingSimilaritySearch2018}.
We refer the reader to the original paper for an in-depth presentation of the technique and the datasets.

\begin{figure}
  \centering
  \includegraphics[trim=0 0 0 0, clip,width=\linewidth]{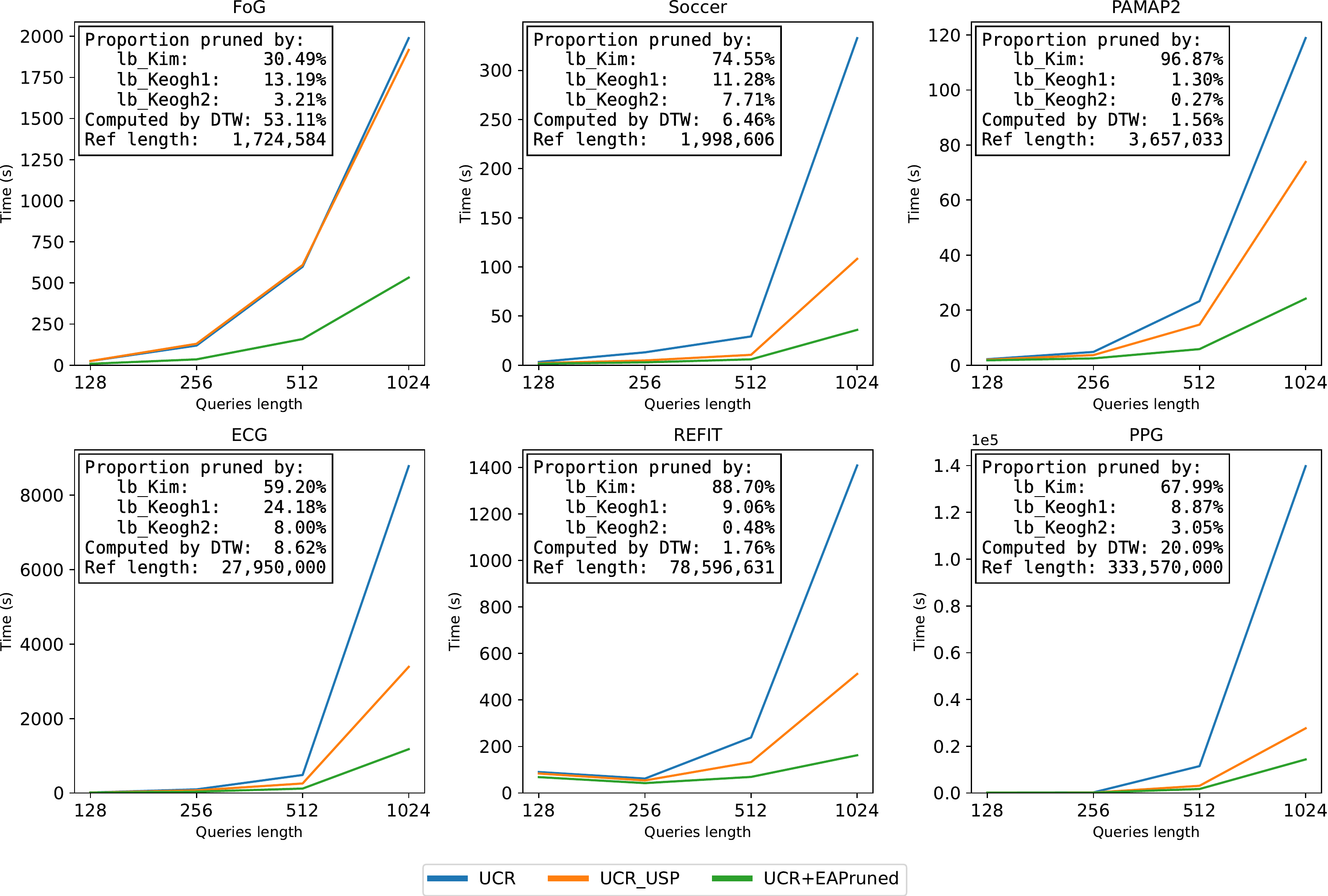}
  \vspace*{-8pt}\caption{\label{fig:expss}
  Comparison of subsequence search time, in second, between the UCR Suite, the UCR-USP Suite,
  and the UCR Suite using EAPruned, when increasing the length of the queries.
  }
\end{figure}

The results are presented in Figure~\ref{fig:expss}, in seconds.
For each dataset, we show the length of the reference series,
and how much DTW computations are saved using three increasingly tighter lower bounds.
The version using our EAPruned implementation is always the fastest.
Moreover, it scales better with long queries.
Overall, the UCR Suite took 4153309 seconds (48 days, 1h:41m49s) to complete,
the UCR-USP Suite took 963251 seconds (11 days, 3h34m11s),
and UCR+EAPruned took 473150 seconds (5 days, 11h25m50s),
providing the fastest known implementation for this task.

\section{\label{sec:5}Conclusion}
EAPruned is an efficient  algorithm for computing elastic distances
that efficiently integrates pruning and early abandoning.
We implemented EAPruned for six key elastic distances
used by state-of-the-art ensemble classifiers,
and compared the timings with existing techniques.

We show experimentally, using the standard UCR archive,
that EAPruned supports the fastest known NN classifiers.
Not only is EAPruned alone competitive against other techniques using lower bounds,
it further benefits from them.
We also show that pruning alone can be beneficial for some distances, allowing the algorithm to be applied productively even when early abandoning is not applicable.
Caution is advised, however, as the overheads of pruning may exceed the benefits in some cases.
Finally, we show that our algorithm can be successfully applied to other instances of similarity search,
like sub-sequence search where its application leads to the fastest known tool in its class.

In light of these results,
we encourage researchers to keep developing lower bounds, the two technique being complementary.
We also encourage practitioners to use our algorithm.
We make the latter easy by releasing the Tempo library \citep{TEMPO},
providing our C++ implementations with Python/Numpy bindings
under the permissive BSD-3 license.

Our next step will be to implement our algorithm for the multivariate case.
We also plan to fit some ensemble classifiers such as Proximity Forest or TSChief
with our distances, expecting significant speed up.
%A formal study of the time complexity of EAPruned is left for further research.

\bibliographystyle{plain}
\bibliography{biblio}

\end{document}